\documentclass[lettersize,journal]{IEEEtran}
\usepackage{cite}
\usepackage{graphicx}
\usepackage{caption}
\usepackage{multirow}
\usepackage{amsmath,amssymb,amsfonts}
\usepackage{amsthm}
\usepackage{mathrsfs}
\usepackage{textcomp}
\usepackage{manyfoot}
\usepackage{booktabs}
\usepackage{algorithm}
\usepackage{algorithmicx}
\usepackage{algpseudocode}
\usepackage{listings}
\usepackage{pgf-pie}
\usepackage{tikz}
\usepackage{float}
\usepackage{tabularx}
\usepackage{enumitem} 
\usepackage{lettrine} 
\usepackage{stfloats}
\usepackage{subcaption}
\usepackage{rotating}
\usepackage{array} 
\usepackage{xcolor} 
\usepackage{graphicx}
\usepackage{latexsym}
\usepackage{amsmath,epsfig}
\usepackage{epstopdf}

\usepackage{epsfig}
\usepackage{amsmath}
\usepackage{amssymb}
\usepackage{mhsetup}
\usepackage{mathtools}
\usepackage{xfrac}
\usepackage{epsfig}
\usepackage{array}
\usepackage{rotating}
\usepackage[numbers]{natbib}
\usepackage{pifont}   
\usepackage{adjustbox}
\usepackage{lscape}      
\usepackage{pdflscape}

\definecolor{darkgreen}{rgb}{0.0, 0.7, 0.0}  

\usepackage[colorlinks=true, linkcolor=blue, citecolor=darkgreen, urlcolor=blue]{hyperref}

\begin{document}

\title{TinyEcoWeedNet: Edge Efficient Real-Time Aerial Agricultural Weed Detection}

\author{
    Omar H. Khater, Abdul Jabbar Siddiqui*
    \thanks{O. H. Khater is with the Computer Engineering Department, King Fahd University of Petroleum and Minerals (KFUPM), Dhahran 31261, Saudi Arabia (e-mail: g202313250@kfupm.edu.sa).}%
    \thanks{*A. J. Siddiqui (Corresponding Author, Co-first author) is with the SDAIA-KFUPM Joint Research Center on Artificial Intelligence, IRC for Intelligent Secure Systems and Department of Computer Engineering, KFUPM (e-mail: abduljabbar.siddiqui@kfupm.edu.sa).}%
}

\author{
    Omar H. Khater*, Abdul Jabbar Siddiqui$^{* \ddagger}$, Aiman El-Maleh, M. Shamim Hossain$^\ddagger$%
    \thanks{*O. H. Khater is with the Computer Engineering Department, King Fahd University of Petroleum and Minerals (KFUPM), Dhahran 31261, Saudi Arabia (e-mail: g202313250@kfupm.edu.sa).}%
    \thanks{*$\ddagger$ A. J. Siddiqui (Corresponding Author, Co-first author) is with the SDAIA-KFUPM Joint Research Center on Artificial Intelligence, Center for Intelligent Secure Systems and the Department of Computer Engineering, KFUPM (e-mail: abduljabbar.siddiqui@kfupm.edu.sa).}%
  \thanks{A. H. El-Maleh is with the Center for Intelligent Secure Systems and the Department of Computer Engineering, King Fahd University of Petroleum \& Minerals, Dhahran 31261, Saudi Arabia (e-mail: aimane@kfupm.edu.sa)} 
  \thanks{$\ddagger$ M. Shamim Hossain 
  is with the Research Chair of Pervasive and Mobile Computing, Department of Software Engineering, College of Computer and Information Sciences, King Saud University, Riyadh 12372, Saudi Arabia (e-mail: mshossain@ksu.edu.sa)}
}

\maketitle

\begin{abstract}
Deploying efficient deep models for real-world applications in agriculture is challenging because the computations and available resources are limited on edge devices. This work proposes a strategy for deploying a compressed version of EcoWeedNet. This efficient deep learning model employs structured channel pruning and quantization-aware training (QAT), while accelerating computations using NVIDIA's TensorRT framework on the Jetson Orin Nano. The proposed approach's critical modules are pruned while preserving their consistency in the pruned architecture, compared with the pruned state-of-the-art YOLO models (YOLO11n and YOLO12n), with performance tested over two benchmarking datasets: an aerial soybean dataset and the benchmarking dataset of CottonWeedDet12.  The inherent complexity of structured pruning for EcoWeedNet and YOLO models arises due to their sophisticated architectural features, that is, residual shortcuts, complex attention mechanisms, concatenations, and CSP blocks, which make it highly challenging to maintain model consistency with structured pruning. Despite all these, structured pruning effectively reduces model parameters by up to 68.5\%, and decreases computational load up to 3.2 GFLOPs. Accelerated inference speed up to 184 FPS with an 11.6\% pruning ratio at lower precision (FP16), improving over 28.7\% compared to Baseline EcoWeedNet at FP16. Our pruned EcoWeedNet model with a pruning ratio of 39.5\% even outperformed YOLO11n and YOLO12n models with significantly lesser pruning ratios of about 20\%, with the precision of 83.7\%, a recall of 77.5\%, and a mAP50 of 85.9\% on the CottonWeedDet12 dataset. These observations validate that the compressed EcoWeedNet is optimal and efficient for precision agriculture.
\end{abstract}

\begin{IEEEkeywords}
Compressed EcoWeedNet, Structured Pruning, Quantization, TensorRT
\end{IEEEkeywords}

\section{Introduction}

Channel pruning approaches can reduce the model's size by eliminating less informative channels, thereby decreasing the neural network's complexity. Channel pruning can be done during the training \cite{saikumar2024drive} or after the training (Post-pruning) \cite{mallya2018packnet}. Pruning during training can be accomplished by selecting the epoch at which pruning takes place, allowing the model to refine and adapt to the reduced architecture. The primary advantage of pruning during training is restoring the model's performance. However, it necessitates additional fine-tuning and computational resources, which can be provided during the training of small networks. In contrast, it is unlikely to be a suitable solution for large models, such as large language models (LLMs). On the other hand, post-pruning eliminates the less informative channels after model training, offering a more straightforward implementation; however, it may cause a notable drop in performance compared to pruning during training. This approach is primarily used in large networks; fine-tuning after pruning is computationally costly. Both techniques enhance the deployment of deep learning models on edge devices by making it lighter and reducing the inference time.

\begin{figure}[h]
    \centering
    \setlength{\fboxsep}{0pt} 
    \setlength{\fboxrule}{0pt} 
    \includegraphics[width=\columnwidth]{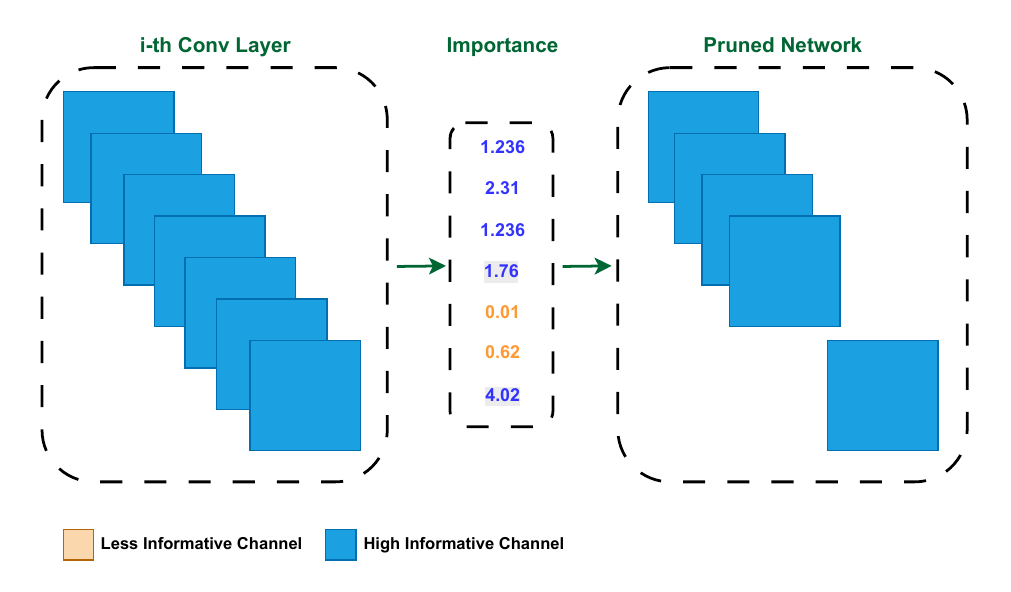}
    \caption{Channel pruning by removing less-informative channels.}
    \label{fig: Pruning_Approach}
\end{figure}

Quantization can be considered a challenging technique utilized for model compression, which reduces the precision of the weights and activations from a 32-bit floating-point representation to a low-bit representation, such as 16-bit floats or 8-bit integers \cite{jacob2018quantization}. The effect of this reduction on the model's memory footprint, which, by nature, accelerates the model inference. The quantization has two major types: the first one is quantization-aware training (QAT), and the other one is called post-training quantization (PTQ) \cite{10536014}. QAT is simulated with low precision during the training process, allowing the model to adapt to the noise introduced by lower precision. However, training requires more computational resources. On the other hand, PTQ is applied to the trained model. However, it is less resource-intensive, which can lead to a significant drop in accuracy.

Integrating the channel pruning and the QAT offers significant potential in model compression \cite{9367271}\cite{kim2021pqk}. While channel pruning reduces the number of channels, quantization decreases the precision of the parameters, resulting in a lightweight model. This integration significantly improves model inference time, especially on NVIDIA TensorRT \cite{10965714}, and reduces the model size while maintaining almost the same performance. Integrating these two approaches can be the optimal choice regarding model compression while retaining the performance deployed on edge devices.

Our work makes multiple significant contributions to solving research problems, substantially advancing the field of automated weed detection using deep learning methods for precision agriculture. A summary of such contributions emphasizes the necessity of more computationally efficient weed detection models that strike a balance between computational efficiency, detection accuracy, and applicability, particularly for devices with limited constrained resources. Our contributions are as follows:

\begin{itemize}

    \item \textbf{Channel Pruning Optimization:} This study pioneers the application of structured pruning to EcoWeedNet \citep{khater2025ecoweednetlightweightautomatedweed}, as well as to the latest members of the YOLO family, YOLO11n \cite{ultralytics2024yolo11} and YOLO12n \cite{ultralytics2025yolov12}. To the best of our knowledge, this study is the first to investigate channel pruning on the YOLO11 and YOLO12 models.

    \item \textbf{Impact of Quantization:} To enhance the efficiency, we integrated quantization-aware training (QAT), allowing the models to learn with reduced precision and produce a compact, deployment-ready architecture without loss of accuracy.

    \item \textbf{Integration of Pruning and Quantization:} The paper explores the integration between structured pruning and QAT approaches on the model's performance.

    \item \textbf{Conducted a comprehensive memory and GPU efficiency analysis:} to evaluate the deployment suitability of the models on the NVIDIA Jetson Orin Nano kit.

    \item \textbf{Comprehensive Validation Approaches:} Validation involved a comprehensive evaluation on two real-world datasets, CottonWeedDet12 and an aerial soybean dataset, and was deployed on the TensorRT engine to confirm EcoWeedNet's applicability in real-world scenarios for the accurate and efficient detection of weeds.

\end{itemize}


\section{Related Works}

Pruning is primarily used to remove unnecessary parameters or channels from the architecture, making it lighter, faster, and more suitable for deployment on resource-constrained devices. The authors in \cite{li2022revisiting} proposed a random pruning approach that challenged the other techniques, which relied on heuristics, such as L1/L2 norms, Taylor expansion, or geometric median. The authors suggested that randomly selecting the pruning ratios and configurations can achieve competitive results compared to the other technique, which relied on complex heuristics. The proposed method was validated across two scenarios. 

The first is pruning pre-trained networks, which is applied by identifying the channels based on random ratios and aligning the feature map before fine-tuning. On the other hand, when pruning from scratch, the authors used parallel sampling of sub-networks to generate compressed models. The random pruning effectiveness can be shown in the results. The technique achieved a 50\% FLOPs reduction on ResNet-50 and scored a top-1 error of 25.22\%, while the Adapt-DCP achieved 24.85\%. The random pruning after the fine-tuning matched the performance of sophisticated methods, such as HRank and AutoPruner, on CIFAR-10. This work emphasized the effectiveness of random pruning as a practical method, offering a simpler framework with reduced computational cost. 

Conversely, the ABCPruner introduced in \cite{lin2020channel} is a highly structured and automated approach. The ABCPruner utilized the Artificial Bee Colony (ABC) algorithm to prune the channels. ABCPruner applied the best channel setup by constraining the search by setting retention percentages in advance, which keeps the network flexible and reduces the computational cost. The authors validated the ABCPruner on the CIFAR-10 dataset and compressed the FLOPs of VGGNet-16 by 73.68\% and the parameter 88.68\%, with a slight improvement in the accuracy by 0.06\%. The ABCPruner was implemented on a larger ILSVRC-2012 dataset and achieved a 56.61\% FLOPs reduction on ResNet-50 and maintained the top-1 accuracy of 73.52\%. The proposed framework combined simplicity with high performance, illustrating the efficiency and scalability.

Due to sensitivity concerns, the authors in \cite{mondal2022adaptive} and \cite{yang2022channel} showed a deep exploration. In \cite{mondal2022adaptive}, the authors introduced Global Filter Importance-based Adaptive Pruning (GFI-AP), which is a global metric used to compute the importance of the filter based on normalized class-specific activations. Oppositely, the authors in \cite{yang2022channel} used second-order sensitivity metrics, which utilized the Hessian matrix as guidance in the pruning process. The GFI-AP ranks the filters to apply the pruning based on each filter rank. The proposed technique was used to VGG16 and reduced the model parameters by 78\% while maintaining the accuracy of 94.23\% on the CIFAR-10 dataset. Regarding the other experiment, the GFI-AP reduced ResNet50 FLOPs by 42.67\% with a small accuracy drop by 1\% on the ImageNet dataset. In \cite{zhang2022carrying}, a white-box framework was introduced to evaluate the contribution of each channel. The proposed technique assigns class-wise masks to assign the channel importance dynamically through the training. The less important channels are pruned based on the global voting mechanism, followed by fine-tuning to recover the lost performance. The white-box approach achieved a reduction in the FLOPs on ResNet110 of  65.23\%, with a small improvement in the accuracy by 0.62\% on CIFAR-10. Additionally, scored a 45.6\% FLOPs reduction on ResNet-50 with a tiny drop in top-1 accuracy.

In \cite{liu2023eacp}, the automation and optimization are deeply investigated. The Effective Automatic Channel Pruning (EACP) approach combines k-means++ clustering and an improved Social Group Optimization (SGO) algorithm. EACP reduces the less important filters by clustering similar filters and keeping the most informative ones. EACP reduced ResNet110 FLOPs by 65.7\% and offered a slight improvement in the accuracy by 0.09\% on CIFAR-10. On the other hand, applying on ImageNet, the EACP achieved a 57.8\% FLOPs reduction on ResNet50 with a small accuracy drop by 1.87\%. Similar to the ABCPruner in \cite{lin2020channel}, but with more focus on reducing the redundancy by applying clustering.

In \cite{geng2024complex}, the Complex Hybrid Weighted Pruning (CHWP) algorithm offers a structured filter pruning framework unifying filter norms, inter-filter dissimilarity, and Batch Normalization (BN) parameters into a single scoring metric. By taking supplementary scaling and shift of BN into account, CHWP can better measure a filter's actual contribution, supporting accurate pruning even for layers with dissimilar or identical filter distributions. Evaluated on CIFAR-10 and ImageNet with several different ResNet architectures, CHWP supported an optimal FLOP reduction by 65.8\% on ResNet-110 with an improved accuracy by 0.16\%, and 53.5\% FLOP reduction on ResNet50 with a loss of only 0.6\% Top-1 accuracy, outperforming others like SFP, FPGM, and WHC. In contrast to unstructured pruning, which induces irregular sparsity and non-accelerability with acceleration libraries, CHWP guarantees a model structure for deployability. Its limitation is that it prunes only convolutional layers, and its efficiency will decline when filter scores are densely packed with minimal redundancy.

Comparison between unstructured and structured pruning is introduced in \cite{zhu2025comprehensive} determines the practical benefit of structured pruning. While unstructured pruning can take out single weights and achieve great sparsity, it produces sparse weight matrices incompatible with GPU libraries and provides little real-world acceleration. It also requires special hardware and complicated fine-tuning, which prevents its deployability. By contrast, structured pruning can eliminate entire filters or channels without changing network shape and obtain real accelerations at inference time on hardware. While it may not provide as much compression as unstructured techniques, structured pruning has a better trade-off on efficiency, simplicity, and deployability. The authors believe structural pruning is a hardware-friendly and scalable model compression technique.

Recently, the importance of quantization has become apparent in optimizing neural network architectures for edge devices, which helps the model to become lighter and memory-efficient while keeping competitive accuracy. There are two main techniques of quantization: Quantization-Aware Training (QAT) and Post-training Quantization. Additionally, the integration of the quantization with the pruning and knowledge distillation techniques\cite{9714269} gives more potential to make the model more efficient for resource-constrained devices.

In \cite{liu2024mpq}, the authors proposed ultra-low mixed-precision quantization employed in the YOLOv5 network. To ensure maximum efficiency, they applied 1-bit precision to the backbone, and to maintain the detection accuracy, they applied 4-bit precision to the Head. The mixed-precision method is complemented by Progressive Network Quantization (PNQ), which gradually quantizes each layer to ensure stable convergence. In the same way, QATFP-YOLO introduced in \cite{idama2024qatfp} utilized QAT to apply low-precision operations during training, which helped to maintain the balance between compression and accuracy of the detection. On the other hand, MPQ-YOLO used 1-bit quantization to push the limits. In QATFP-YOLO, the filter pruning was integrated with 8-bit precision to improve the model performance further. The results for these two approaches illustrated their efficiency. MPQ-YOLO outperformed techniques like Q-YOLO and DoReFa-Net, achieved a reduction of 16.3× in computational complexity, and showed significant improvements in the accuracy of the COCO dataset. QATFP-YOLO achieved 4x better than YOLO-Lite by achieving an inference speed of 88 FPS on a Google Pixel 4 smartphone, showing the potential of combining pruning with QAT for edge devices. Based on the previous results, the trade-offs between the aggressive quantization strategies, such as MPQ-YOLO, and more balanced techniques, QATFP-YOLO, focus on making deployment easier.

The TinyissimoYOLO model, which integrated 8-bit quantization with scalable configurations suitable for deployment on ultra-low-power micro-controllers such as GAP9, was introduced in \cite{moosmann2023flexible}. The NE16 accelerator on GAP9 offered better latency and energy efficiency than the MAX78000, emphasizing the effectiveness of the quantization optimized for resource-constrained devices. On the other hand, the authors of \cite{xue2024yolo} proposed EL-YOLO, which includes TensorRT FP16 quantization, utilized for small-object detection in UAV imagery. The authors showed the impact of combining lightweight architectural components, such as SPD-Conv and CSL-MHSA, with quantization, achieving a 45\% reduction regarding the number of the parameters and an 88\% reduction in FLOPs compared with YOLOv8s, additionally achieved inference speed of 35 FPS. The PG-YOLO in \cite{dong2022pg} confirmed the effectiveness of combining the quantization with R-pruning, and Ghost modules. The proper model achieved an 8.75× compression of YOLOv5s with a small drop in the accuracy by 0.1\%. In the same way, the APDNet proposed by the authors of \cite{zhang2024robust} integrated 8-bit quantization with teach-assistant distillation (TAD), emphasizing the effectiveness of knowledge transfer from a high-capacity teacher model to a lightweight student model. Both approaches showed the impact of integrating quantization and pruning or distillation, which can develop efficient models without sacrificing performance. PG-YOLO is used in industrial IoT applications and scored an inference speed of 30.3 FPS on the Jetson TX2. On the other hand, APDNet is utilized for aerial person detection and achieved a mAP@0.5 of 94.86\% on the Jetson Xavier NX. So, the effectiveness of quantization in industrial and aerial applications can be confirmed.

In \cite{su2024yolic}, the authors introduced YOLIC, which integrated Quantization-aware training in the architecture. The enhanced model reduced the computational complexity while leveraging QAT for further optimization by replacing traditional bounding box regression with Cells of Interest (CoIs). The model achieved 40.06 FPS on a Raspberry Pi 4B. YOLIC showed an outstanding performance compared to YOLOv5-N and YOLOv8-N. The feature aggregation and detection accuracy, especially for small objects, are enhanced by integrating CSL-MHSA in the EL-YOLO architecture.


\begin{table*}[t]
    \centering
    \caption{Summary of Related Works and their Limitations}
    \label{tab:related_works_pruning_quant}
    \resizebox{\textwidth}{!}{%
    \begin{tabular}{c|c|c}
        \hline
        \textbf{Reference} & \textbf{Method} & \textbf{Limitations} \\
        \hline
        \cite{li2022revisiting} & Random channel pruning without heuristics & Simpler framework, but lacks structured guarantees and stability. \\
        \hline
        \cite{lin2020channel} & ABCPruner using Artificial Bee Colony algorithm & Requires fixed retention ratios; flexibility limited to predefined settings. \\
        \hline
        \cite{mondal2022adaptive} & GFI-AP: Global filter importance via class-specific activations & Requires class-specific normalization; moderate accuracy drop on large datasets. \\
        \hline
        \cite{yang2022channel} & Pruning using second-order sensitivity (Hessian) & Computationally expensive due to Hessian matrix calculation. \\
        \hline
        \cite{zhang2022carrying} & White-box pruning with class-wise channel masks and voting & Relies on fine-tuning and mask assignments; adds training complexity. \\
        \hline
        \cite{liu2023eacp} & EACP: Clustering + Social Group Optimization & Effective but depends on filter similarity; performance may degrade on noisy data. \\
        \hline
        \cite{geng2024complex} & CHWP: Combines filter norms, BN stats, and dissimilarity & Limited to convolutional layers; less effective when redundancy is low. \\
        \hline
        \cite{zhu2025comprehensive} & Structured vs. unstructured pruning comparison & Unstructured pruning incompatible with hardware acceleration; structured has lower compression. \\
        \hline
        \cite{liu2024mpq} & MPQ-YOLO: Mixed-precision quantization with PNQ & 1-bit compression may harm generalization; requires careful balancing. \\
        \hline
        \cite{idama2024qatfp} & QATFP-YOLO: QAT + 8-bit pruning integration & Dependent on training hardware; may limit portability. \\
        \hline
        \cite{moosmann2023flexible} & TinyissimoYOLO with 8-bit quant. on GAP9 microcontrollers & Targeted for GAP9; generalizability to other hardware not shown. \\
        \hline
        \cite{xue2024yolo} & EL-YOLO: TensorRT FP16 + lightweight modules & Performance may drop in large-scale scenarios due to small-object focus. \\
        \hline
        \cite{dong2022pg} & PG-YOLO: R-pruning + quantization + Ghost modules & Slight accuracy loss (0.1\%) despite high compression (8.75×). \\
        \hline
        \cite{zhang2024robust} & APDNet: 8-bit quantization + teacher-student distillation & Requires large teacher models; distillation setup adds complexity. \\
        \hline
        \cite{su2024yolic} & YOLIC: QAT + CoI regression + CSL-MHSA & Designed for Raspberry Pi; scalability to more demanding tasks untested. \\
        \hline
    \end{tabular}%
    }
\end{table*}

\section{Proposed Method}

Our proposed methodology introduces an end-to-end optimized framework for improving computational efficiency and enhancing the applicability of deployment on edge devices for the EcoWeedNet model, which is used for weed detection. This section first outlines the structured pruning approach used alone, explicitly engineered to eliminate less informative channels with minimal loss of detection accuracy while maintaining consistency in the architecture. Then, quantization-aware training (QAT) is further integrated with NVIDIA TensorRT to improve the model's adaptability to low-precision operations, resulting in significant improvements in inference speed and efficiency. Finally, the discussion of synergistic integration between structured pruning and QAT, deployed on NVIDIA Jetson Orin Nano, illustrates the impact on detection accuracy and inference speed.

\subsection{Structured Pruning}

This work proposes enhancing the computational efficiency of the EcoWeedNet \cite{khater2025ecoweednetlightweightautomatedweed}, YOLO11n \cite{ultralytics2024yolo11}, and YOLO12n \cite{ultralytics2025yolov12} architectures through channel pruning, thereby removing less informative channels from the feature map. The pruning procedure targeted every convolutional layer in the network, except for the Detect layer, which is utilized for multi-scale output predictions and is ideally kept in its original dimensional consistency for precise detection.

The pruning approach involves calculating the  \(\ell_1\)-norm of filter weights in the feature map. The importance score $I_c$ for a specific output channel $c$ in a convolutional layer is determined as:

\begin{equation}
I_c = \sum_{i,j,k} |W_c(i,j,k)|
\end{equation}

where $W_c(i,j,k)$ are the weights corresponding to the $c$-th output channel. Channels with the lowest  \(\ell_1\)-norm values were considered less informative and removed, as shown in Figure~\ref{fig: Pruning_Approach}, reducing the number of parameters and FLOPs.

A primary contribution of this work is overcoming the inherent challenges associated with pruning the following complex modules within EcoWeedNet and benchmarking them with the pruned versions of YOLO11n and YOLO12n:

\textbf{1. Conv Layers:} These are considered foundational blocks of the architecture, and they have a convolutional layer followed by batch normalization and an activation function within each block. Pruning these layers is straightforward but crucial, as they exist throughout the network. The \(\ell_1\)-Norm importance scores were applied to these layers, and batch norm layers were updated to preserve normalization behavior post-pruning.

\textbf{2. C3K2 Module:} The C3K2 module includes a C2f with multiple nested bottleneck layers. It uses stacked CSP-like blocks and shortcuts within a deep residual structure. The initial split projection cv1 and final aggregation cv2 layers were pruned, along with the internal blocks, to optimize this module. It included coordinating pruning indices among branches to preserve dimensional concatenation consistency and perform skip paths.

\textbf{3. C2PSA Module:} The C2PSA module combines a convolutional splitter, multiple parallel PSA attention blocks, and a final projection. The challenge was to prune the PSA feed-forward pathway (the second branch in the split), which carries the self-attended features. The pruning was done on both cv1 and cv2 while maintaining the balance and compatibility across the split-merge operation.

\textbf{4. SPPF (Spatial Pyramid Pooling - Fast):} The SPPF block aggregates features from multiple receptive fields via repeated max-pooling and channel-wise concatenation. It uses an initial projection layer, cv1, followed by three sequential max pooling layers and a final projection, cv2. Pruning this module required modifying the convolutional layers cv1 and cv2. These guarantee compatibility in the concatenated output dimensions across all pyramid branches.

\textbf{5. SPAB (Swift Parameter-Free Attention Block):} (EcoWeedNet only) SPAB contains three convolutional layers with intermediate residual-based modulation. Pruning this block required isolating the convolutional paths c1\_r, c2\_r, and c3\_r while preserving the residual connection and attention modulation computed as ($\sigma(out3) - 0.5$). Special care was taken to prune only the filters and batch norms of the convolutional layers without altering the residual path. Pruning was performed independently across the three layers, and attention computation was kept channel-consistent.

\textbf{6. A2C2f Module:}(YOLO12n only) A2C2f module is composed of convolutional projections, sequential area-attention blocks (ABlock/C3k), and a residual scale parameter. Pruning was selectively implemented on the channels of the first projection (cv1), sequential attention blocks, and the last projection (cv2) with dimensional consistency and usability of the residual connections regulated by the learnable parameter (gamma).

These modules significantly enhance the performance of EcoWeedNet but complicate the channel pruning approach on EcoWeedNet, YOLO11n, and YOLO12n. This strategy effectively reduced model complexity without significant performance loss, later validated by experimental results. The Detect layer was kept completely so that stable and accurate bounding box predictions can be made across different feature map scales.

\subsection{Quantization-Aware-training}

We begin by replacing every standard Conv layer in the EcoWeedNet network with a variant that wraps an NVIDIA TensorQuantizer.

With the quantizers switched to “observe–only” mode, the network processes a small batch of images. Each quantizer builds a histogram of its activations and selects amax and scale values that map those activations into the integer range [-128, 127] for hardware compatibility. Once the optimal scales are fixed, the quantizers are re-enabled, allowing every subsequent forward pass to apply 8-bit precision.

During the forward pass, each quantizer clamps activations as if they were stored in 8-bit integers, yet the underlying weights remain full-precision floating point, so back-propagation behaves normally. Training then continues exactly as usual, but now every forward and backward pass carries the noise introduced by the simulated 8-bit arithmetic. Because the weights experience quantization error during learning, this phase, known as quantization-aware training (QAT), allows the model to adapt its parameters to low-precision computation. After QAT is complete, the resulting checkpoint is typically deployed in lower precision for inference, ensuring compact storage and faster inference speeds.

\subsection{Integration between Structured Pruning and QAT}

We begin by preparing the network for quantization. All Conv layers are swapped for simulated-quantized layers, and a calibration pass fixes the 8-bit scaling factors, so every forward step now mimics INT8 arithmetic while gradients remain in full precision. Then, channel pruning is triggered at a chosen epoch and pruning ratio, trimming away the least-important channels and copying informative weights into the slimmer architecture. Training then resumes and remains under quantization-aware, allowing the model to simultaneously adapt to its reduced width and the quantization noise during the remaining fine-tuning epochs \cite{10891155}. When training finishes, the final checkpoint is exported to lower precision, resulting in a compact file that can be deployed on the NVIDIA Jetson Orin Nano kit.

\section{Experimental Setup}

\subsection{Dataset}

\begin{figure}[h]
    \centering
    \setlength{\fboxsep}{0pt} 
    \setlength{\fboxrule}{1pt} 
    \fbox{\includegraphics[width=\columnwidth, keepaspectratio]{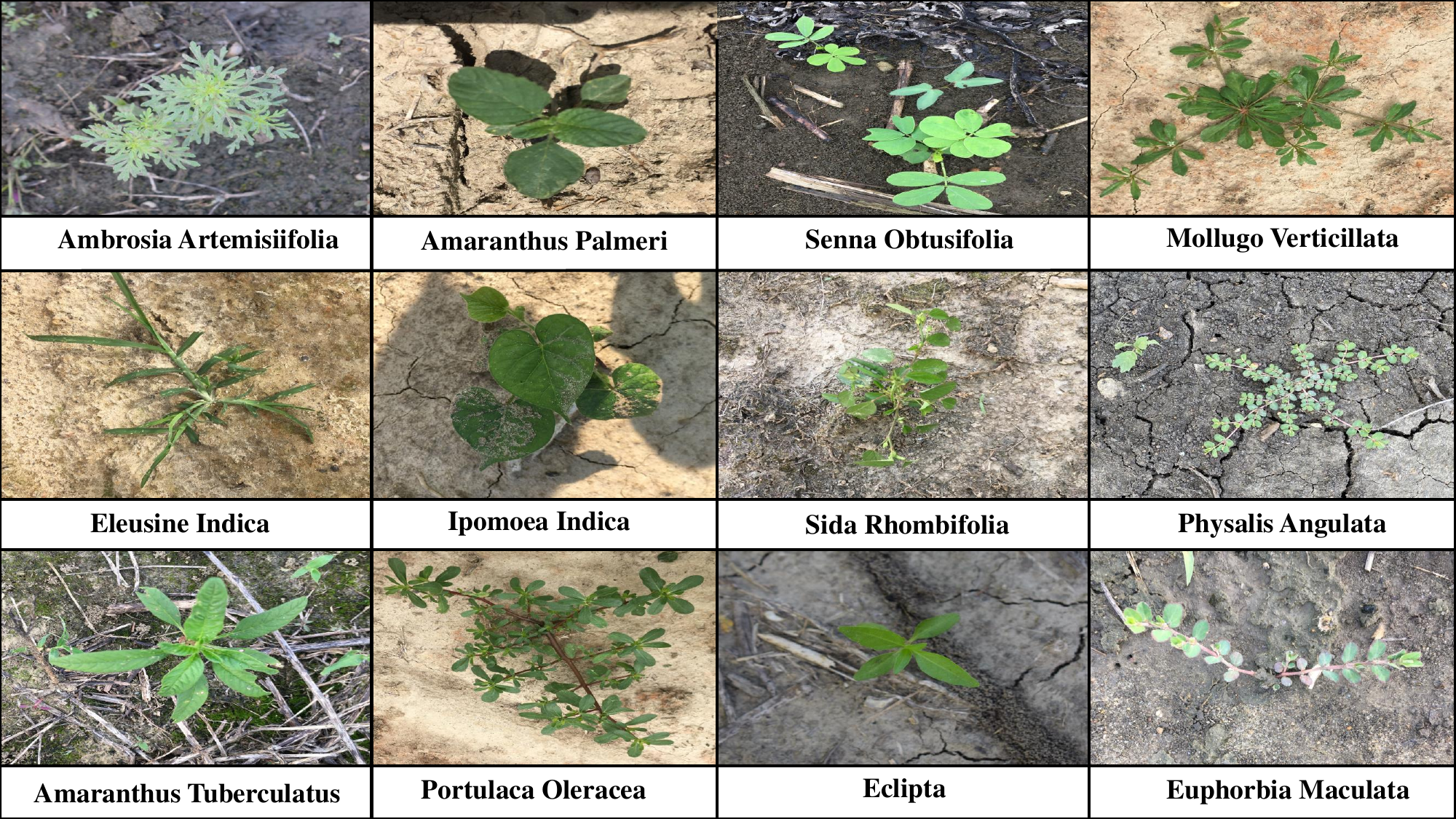}}
    \caption{CottonWeedDet12 Dataset Sample \cite{khater2025ecoweednetlightweightautomatedweed}}
    \label{fig: ground_dataset_samples}
\end{figure}

\begin{figure}[h]
    \centering
    \setlength{\fboxsep}{0pt} 
    \setlength{\fboxrule}{1pt} 
    \fbox{\includegraphics[width=\columnwidth, keepaspectratio]{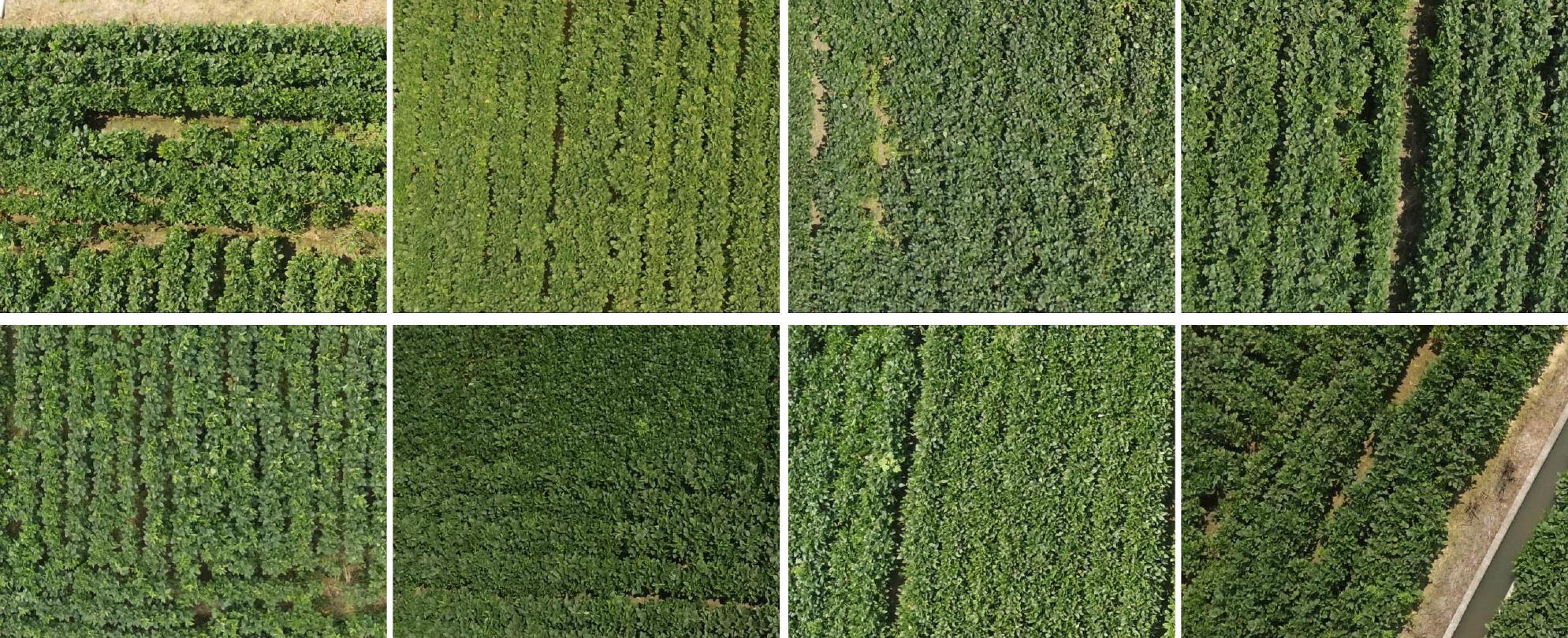}}
    \caption{Aerial Soybeans Dataset Sample}
    \label{fig: drone_dataset_samples}
\end{figure}

In our work, we aimed to utilize two types of datasets. The first is the CottonWeedDet12 dataset, which consists of 12 common weed species found in cotton fields in the southern region of the United States 
\cite{lu2023cottonweeddet12}. This dataset captures close-up images of the weeds from the ground. The second dataset \cite{rehman2024agriweed} focuses on weed detection in a soybean farm, captured from a drone. Providing a strong and varied testing of EcoWeedNet. The CottonWeedDet12 dataset contains 5,648 high-resolution RGB images annotated with 9,370 bounding boxes, making CottonWeedDet12 a perfect, openly available benchmark. The used dataset includes the shadow effect, complex background, and multiple kinds per image, as shown in Figure~\ref{fig: ground_dataset_samples}, emphasizing the real-world scenarios and enhancing the model performance, robustness, and generalization. While the second dataset, as shown in Figure~\ref{fig: drone_dataset_samples}, includes 2,247 images for only one class (weed), each including 15,225 bounding boxes in total.

Both of them were divided into 80\% for training, 10\% for validation, and 10\% for testing. Images were captured in various environmental settings. This significantly contributes to the model's generalization to real-world scenarios. These rich datasets provide a solid foundation for training and testing the proposed model to perform effectively in weed detection tasks.

\subsection{Hardware and Software}

The baseline YOLO11n and YOLO12n models from the Ultralytics library were implemented in PyTorch, exported to ONNX, and subsequently converted to TensorRT engines for deployment. All inference experiments were conducted on an NVIDIA Jetson Orin Nano Developer Kit \cite{10948450}\cite{yolotrt2022}.

\subsection{Evaluation Metrics}

The evaluation metrics used to evaluate the model’s performance are presented below in Table \ref{tab: evaluation_metrics}:

\begin{table}[h]
\centering
\caption{Evaluation metrics}
\label{tab: evaluation_metrics}
\renewcommand{\arraystretch}{2.5}

\resizebox{\columnwidth}{!}{%
\begin{tabular}{c|c|c}
\hline
\textbf{Metric} & \textbf{Formula} & \textbf{Description} \\ \hline
Precision & $\displaystyle \frac{\text{TP}}{\text{TP} + \text{FP}}$ & Correctly predicted positives. \\ \hline
Recall & $\displaystyle \frac{\text{TP}}{\text{TP} + \text{FN}}$ & Actual positives detected. \\ \hline
mAP & $\displaystyle \frac{1}{N} \sum_{i=1}^{N} \text{AP}_i$ & Average precision over all classes. \\ \hline
mAP@50 & -- & mAP at IoU threshold of 0.50. \\ \hline
mAP(50--95) & -- & Mean mAP from IoU 0.50 to 0.95. \\ \hline
Parameters & -- & Total number of trainable weights. \\ \hline
GFLOPs & -- & Giga Floating Point Operations. \\ \hline
Inference Speed (TensorRT) & -- & Inference speed in frames/sec. \\ \hline
\end{tabular}%
}
\end{table}


\section{Results And Analysis}

This section presents and discusses the complete experiments on the aerial soybeans and CottonWeedDet12 datasets after compressing EcoWeedNet, YOLO11n, and YOLO12n. We report the baseline (full-precision) performance and then show how quantization-aware training (QAT) and structured channel pruning affect detection accuracy, model size, and computational cost. Detection accuracy is evaluated with standard metrics (precision, recall, and mAP), while memory consumption is examined to highlight the footprint reduction achieved by each compression stage. We also analyze GPU efficiency through throughput and utilization statistics, measuring end-to-end inference speed, the full round-trip from CPU memory transfer to GPU execution and the return of results to the CPU, to show the actual time consumed. Finally, we evaluate the integrated pipeline for channel pruning and QAT to illustrate the combined gains and trade-offs, demonstrating that the proposed methodology consistently meets the required frames per second (FPS) for real-time deployment on high-speed drones.

\subsection{Results After Pruning}

In this section, we discussed and analyzed the impact of channel pruning on detection accuracy and the size of EcoWeedNet, YOLO11n, and YOLO12n. We conducted intensive experiments on the three models at various channel pruning ratios and epochs. Then, we fine-tuned the models in each of these experiments to recover their detection accuracy.

\subsubsection{Results on Aerial Soybeans Dataset} 

We allowed the network to train almost to convergence at epoch 200, to let the dense models achieve higher precision, recall, and mAP by the end of training, as shown in Table~\ref{tab: EcoWeedNet_performance_SimAM_SPAB_aerial}.

\begin{table}[H]
\centering
\caption{Comparative Performance of the base-line EcoWeedNet, YOLO11n, and YOLO12n.}
\label{tab: EcoWeedNet_performance_SimAM_SPAB_aerial}
\resizebox{\columnwidth}{!}{%
\begin{tabular}{ccccccc}
\hline
\textbf{Model} & \textbf{Precision (\%)} & \textbf{Recall (\%)} & \textbf{mAP50 (\%)} & \textbf{mAP(50-95) (\%)} & \textbf{Parameters} & \textbf{(GFLOPs)} \\
\hline

\textbf{EcoWeedNet\cite{khater2025ecoweednetlightweightautomatedweed}} & 71.4 & 65.7 & 72.5 & 40.8 & 2.78M & 9.3 \\ 
\hline
\textbf{YOLO11n\cite{ultralytics2024yolo11}} & 69.9 & 63.9 & 70.6 & 39.4 & 2.6M & 6.5 \\ 
\hline
\textbf{YOLO12n\cite{ultralytics2025yolov12}} & 69.7 & 62 & 70.3 & 38.1 & 2.6M & 6.5 \\ 
\hline
\end{tabular}%
}
\end{table}

\begin{figure}[ht]
    \centering

    \begin{minipage}{0.5\columnwidth}
        \includegraphics[width=\linewidth]{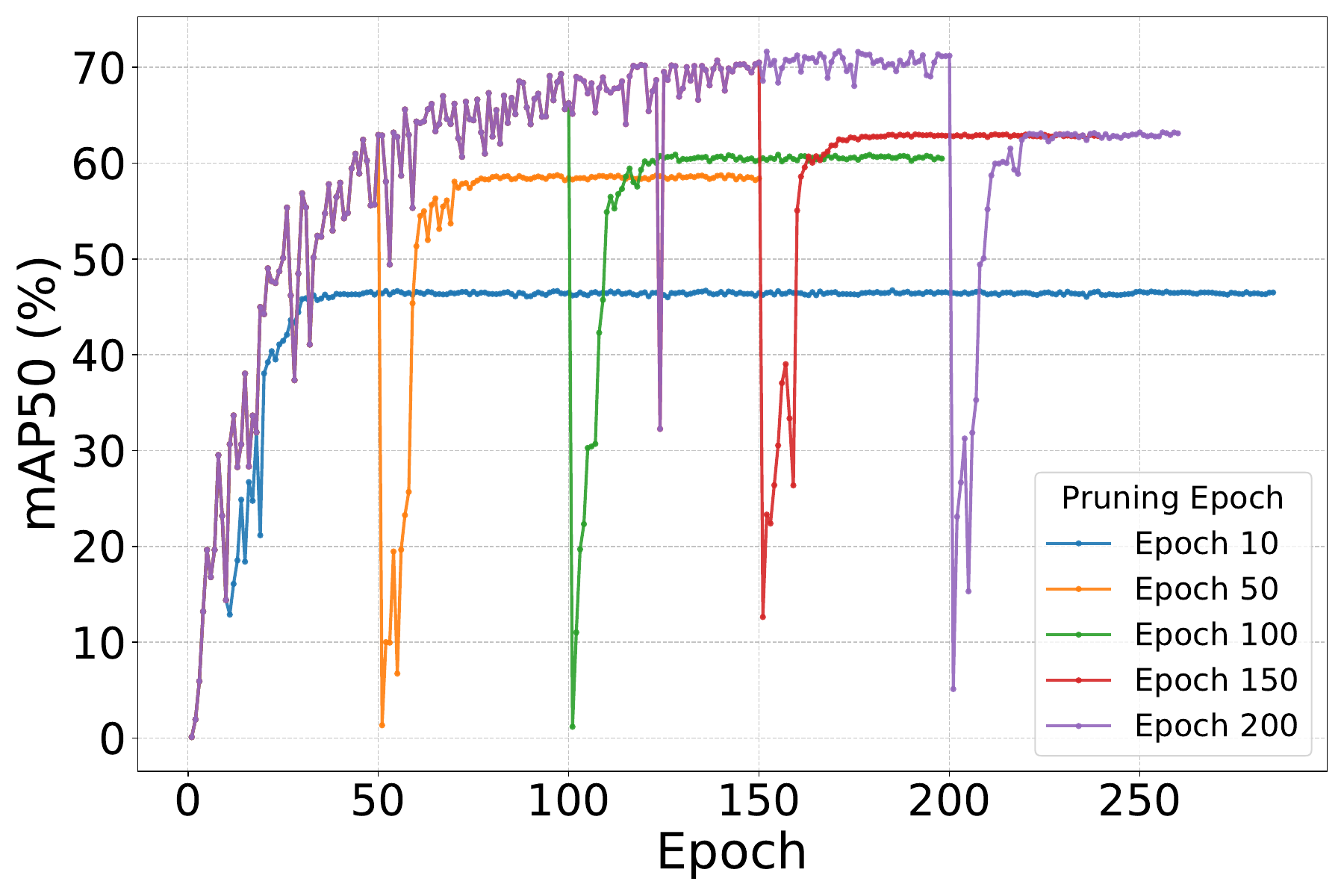}
        \subcaption{Pruning Ratio 11.6\%}
        \label{fig:Eco_mAP_50_10}
    \end{minipage}\hfill
    \begin{minipage}{0.5\columnwidth}
        \includegraphics[width=\linewidth]{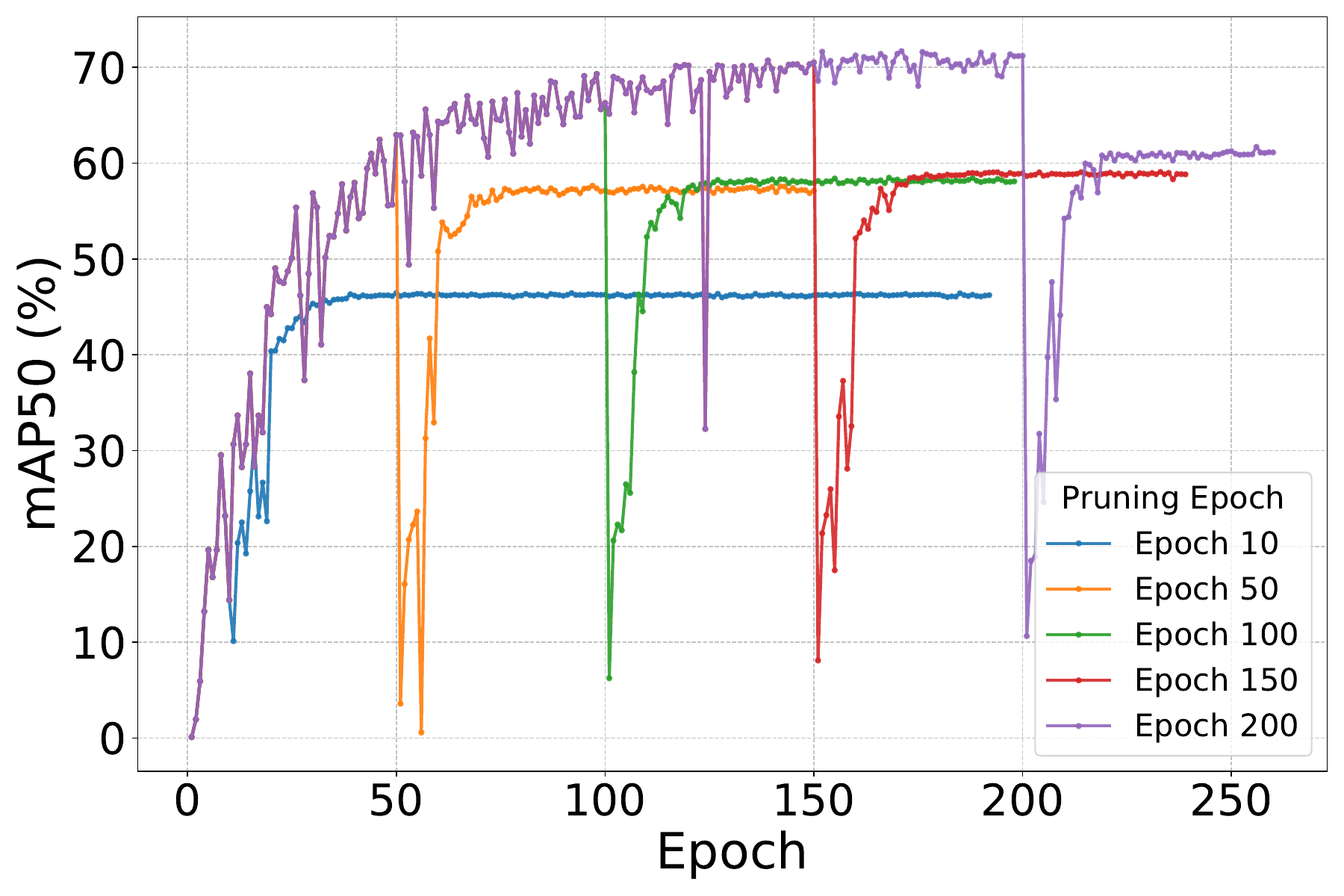}
        \subcaption{Pruning Ratio 21.1\%}
        \label{fig:Eco_mAP_50_20}
    \end{minipage}

    \vspace{0.3cm}

    \begin{minipage}{0.5\columnwidth}
        \includegraphics[width=\linewidth]{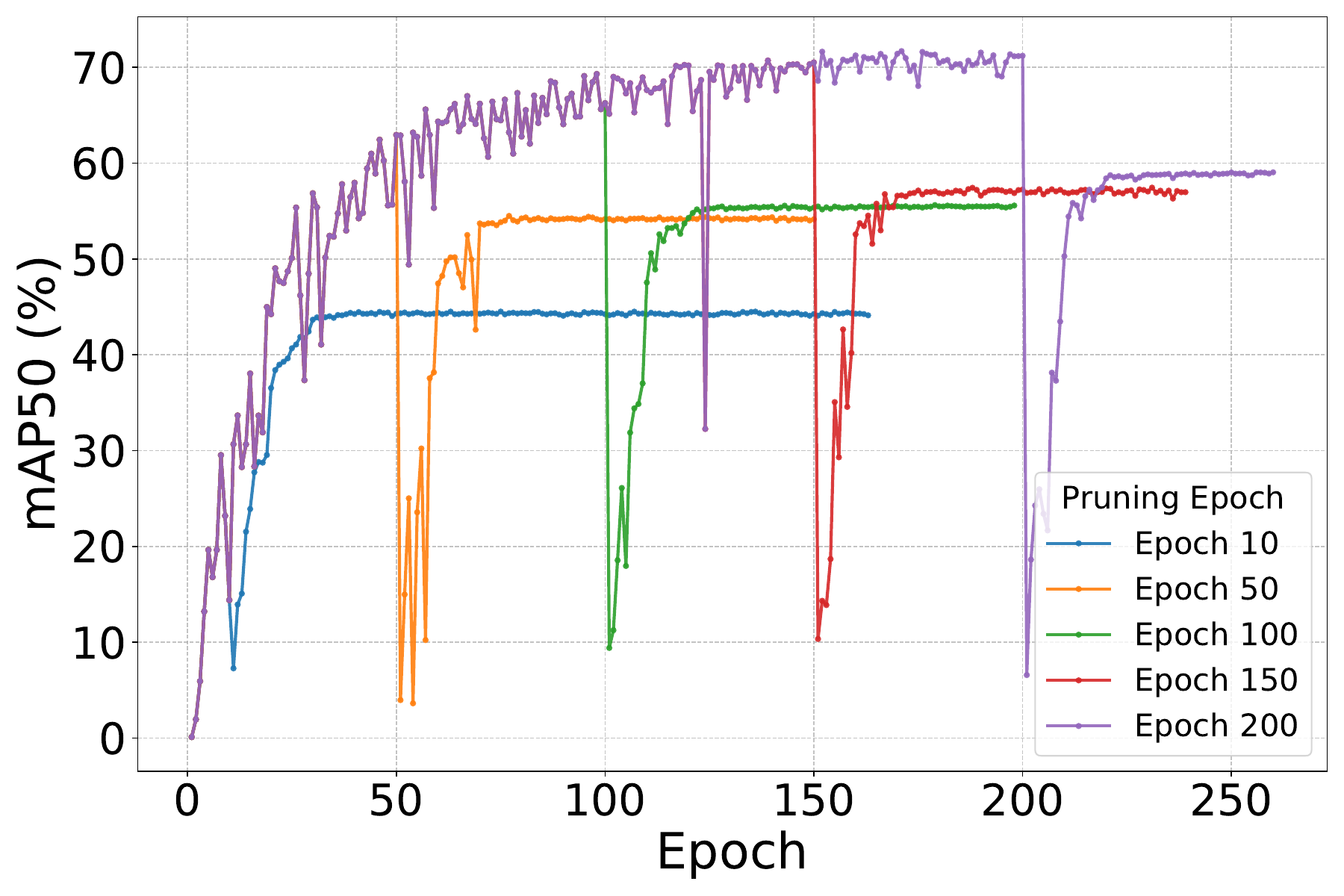}
        \subcaption{Pruning Ratio 30.6\%}
        \label{fig:Eco_mAP_50_30}
    \end{minipage}\hfill
    \begin{minipage}{0.5\columnwidth}
        \includegraphics[width=\linewidth]{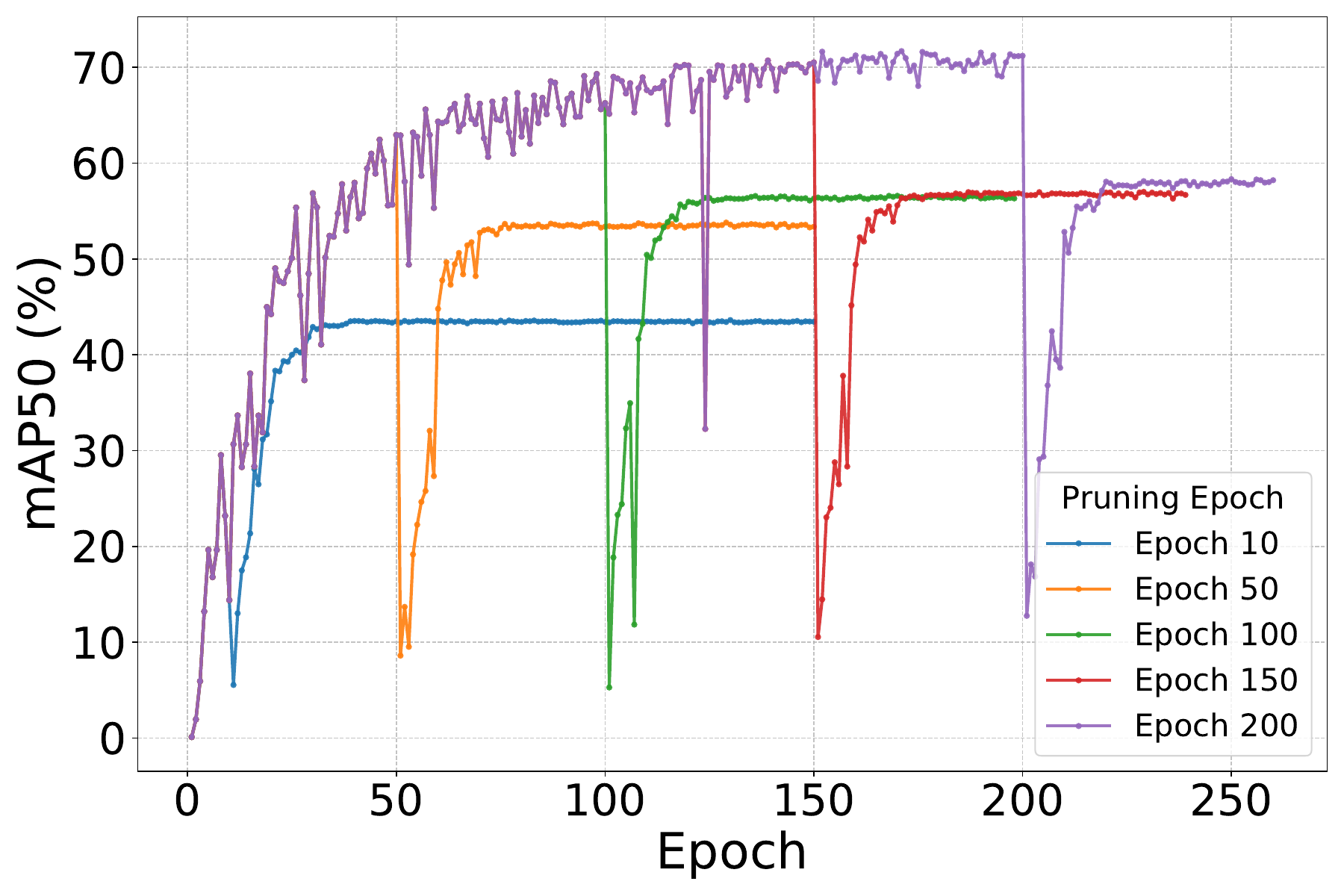}
        \subcaption{Pruning Ratio 39.5\%}
        \label{fig:Eco_mAP_50_40}
    \end{minipage}

    \vspace{0.3cm}

    \begin{minipage}{0.5\columnwidth}
        \includegraphics[width=\linewidth]{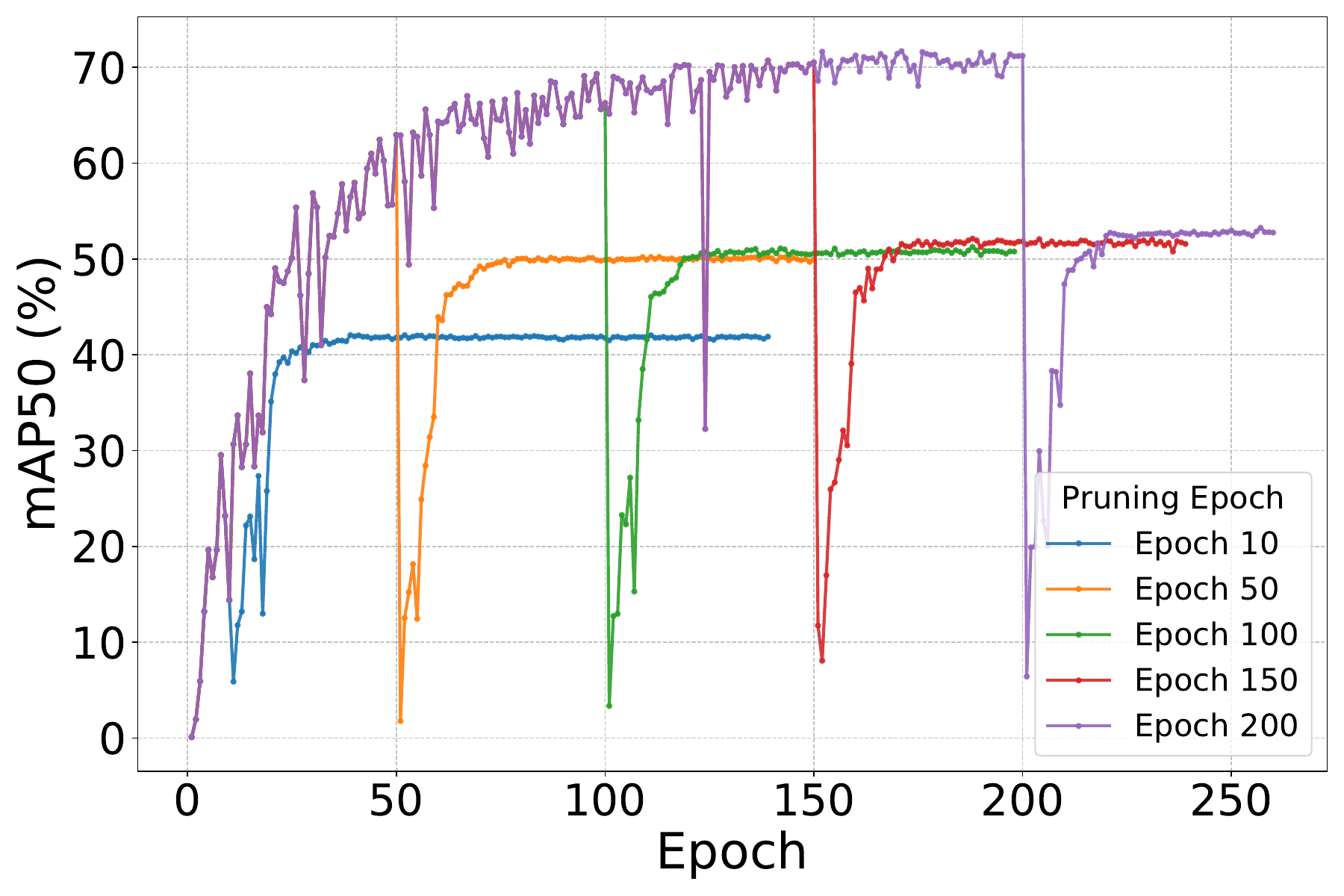}
        \subcaption{Pruning Ratio 55.3\%}
        \label{fig:Eco_mAP_50_55}
    \end{minipage}\hfill
    \begin{minipage}{0.5\columnwidth}
        \includegraphics[width=\linewidth]{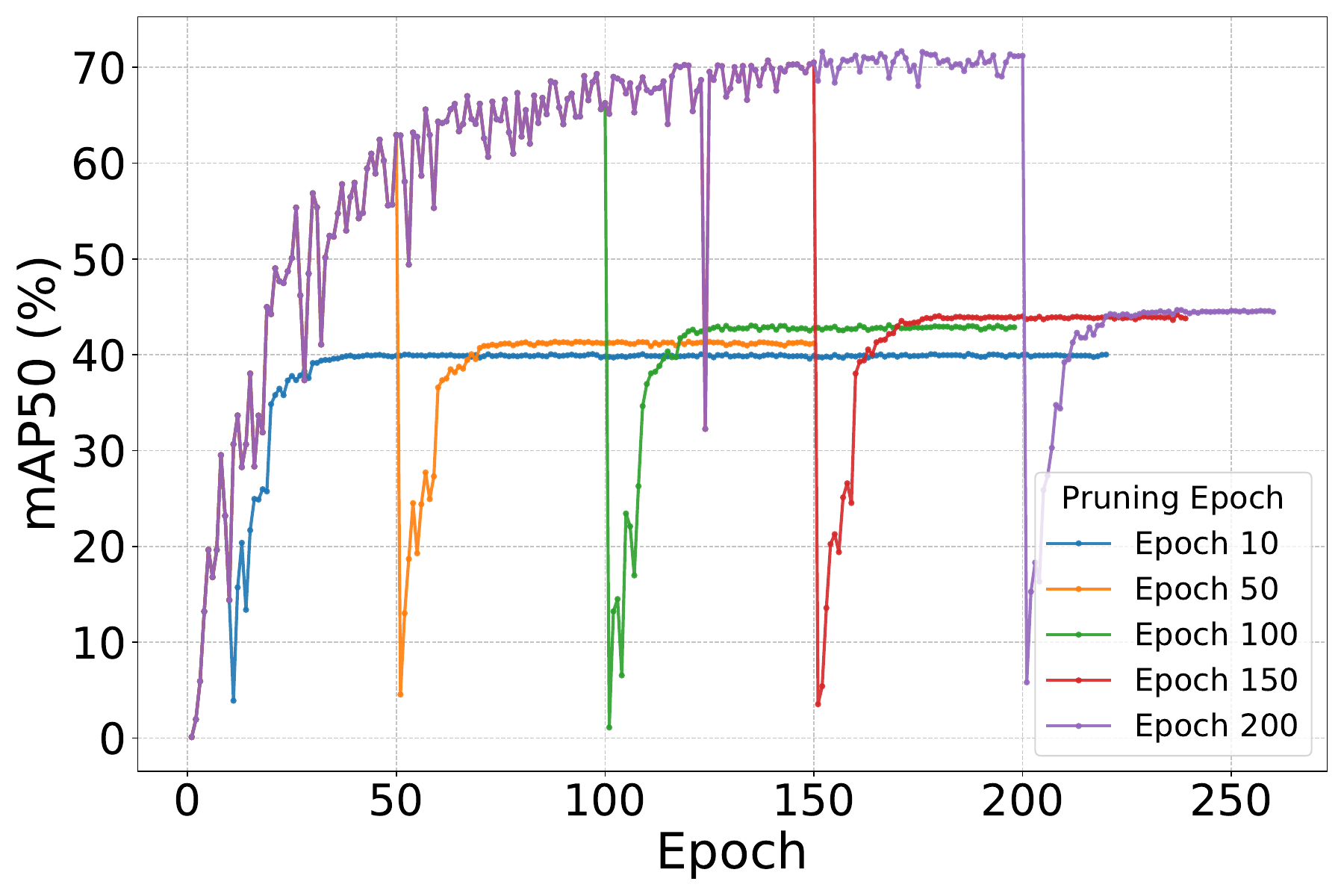}
        \subcaption{Pruning Ratio 68.5\%}
        \label{fig:Eco_mAP_50_68}
    \end{minipage}

    \caption{mAP\textsubscript{50} of EcoWeedNet at different pruning ratios and epochs on the aerial dataset.}
    \label{fig: Eco_mAP_50_Aerial_Combined}
\end{figure}

As shown in Figure~\ref{fig: Eco_mAP_50_Aerial_Combined}, we observed that aggressive pruning, which involves high pruning ratios, significantly degrades detection accuracy in the three models. Additionally, applying pruning later in the training cycle, after the model's performance has stabilized, indicating that it has learned the underlying patterns in the dataset, enhances the model's ability to adapt and recover accuracy effectively.

Pruned EcoWeedNet outperformed both pruned YOLO11n and pruned YOLO12n on aerial soybeans across most pruning ratios, as shown in Table~\ref{tab: Models_Pruned_Results_Aerial}, achieving superior results in mean average precision. Specifically, EcoWeedNet achieved 61.7\% in mAP\textsubscript{50} with a pruning ratio of 21.1\% and 2,196,937 parameters, while YOLO11n attained 60.4\% with an 11\% pruning ratio and 2,305,131 parameters. YOLO12n scored 60.7\% with 2,313,340 parameters and a pruning ratio of 10\%.

Therefore, the best practice in pruning is to let the dense models train until they are nearly converged, then prune them and allow them to recover until the curves flatten again to get the best performance and the lowest degradation, as illustrated in Figure~\ref{fig: Eco_mAP_50_Aerial_Combined}.

\begin{table*}[t]
\centering
\caption{Performance of EcoWeedNet, YOLO11n, and YOLO12n under different channel pruning ratios on aerial soybeans dataset.}
\label{tab: Models_Pruned_Results_Aerial}
\resizebox{\textwidth}{!}{%
\begin{tabular}{c|c|cccc|c|c}
\hline
\textbf{Model} & \textbf{Pruning Ratio (\%)} & \textbf{Precision (\%)} & \textbf{Recall (\%)} & \textbf{mAP50 (\%)} & \textbf{mAP(50--95) (\%)} & \textbf{Parameters} & \textbf{GFLOPs} \\
\hline

\multirow{6}{*}{\rotatebox{90}{EcoWeedNet}}  & 68.5 & 51.4 
 & 43.7 & 44.6 & 21.5 & 876,859 & 3.2  \\[2pt]
 & 55.3 & 54.8 & 49.1 & 53.2 & 26.6 & 1,243,853 & 4.4 \\[2pt]
 & 39.5 & 58.7 & 52.0 & 58.3 & 28.5 & 1,683,879 & 5.8 \\[2pt]
 & 30.6 & 60.4 & 53.1 & 58.8 & 29.7 & 1,931,397 & 6.6 \\[2pt]
 & 21.1 & 61.6 & 54.6 & 61.7 & 31.0 & 2,196,937 & 7.4 \\[2pt]
 & 11.6 & 64.5 & 56.4 & 63.4 & 34.3 & 2,459,176 & 8.7 \\[2pt]
 
\hline

\multirow{6}{*}{\rotatebox{90}{YOLO11n}} & 68.0 & 50.0 & 44.3 & 45.6 & 22.0 & 828,371 & 2.4 \\[2pt]
 & 55.0 & 55.5 & 46.6 & 50.7 & 25.2 & 1,154,797 & 3.3 \\[2pt]
 & 40.0 & 60.0 & 52.8 & 57.6 & 29.2 & 1,564,599 & 4.2 \\[2pt]
 & 30.0 & 60.3 & 52.0 & 57.9 & 29.8 & 1,813,479 & 4.9 \\[2pt]
 & 20.0 & 60.4 & 55.9 & 59.7 & 30.4 & 2,043,929 & 5.3 \\[2pt]
 & 11.0 & 60.6 & 55.3 & 60.4 & 31.4 & 2,305,131 & 6.0 \\[2pt]
 
\hline

\multirow{6}{*}{\rotatebox{90}{YOLO12n}} & 63.3 & 47.5 & 38.2 & 45.9 & 21.6 & 942,611 & 2.6 \\[2pt]
 & 50.6 & 54.4 & 42.4 & 50.5 & 25.5 & 1,267,859 & 3.5 \\[2pt]
 & 37.0 & 57.0 & 48.3 & 57.5 & 29.0 & 1,613,675 & 4.0 \\[2pt]
 & 28.0 & 58.8 & 52.9 & 58.7 & 29.5 & 1,847,694 & 4.8 \\[2pt]
 & 19.3 & 57.9 & 50.2 & 59.8 & 30.3 & 2,072,987 & 5.4 \\[2pt]
 & 10.0 & 61.9 & 51.6 & 60.7 & 31.7 & 2,313,340 & 6.0 \\[2pt]
\hline
\end{tabular}%
}
\end{table*}

\vspace{0.2cm}

\subsubsection{Results on CottonWeedDet12 Dataset}

Based on figure~\ref{fig: Eco_mAP_50_Aerial_Combined}, the optimal pruning behaviour is after the dense model convergence, so we pruned the three models after 100 epochs, which is enough for the convergence using CottonWeedDet12, as shown in Table~\ref{tab: EcoWeedNet_performance_SimAM_SPAB_Ground}.

\begin{table}[H]
\centering
\caption{Comparative Performance of the baseline EcoWeedNet, YOLO11n, YOLO12n, and YOLO4 on CottonWeedDet12 \cite{khater2025ecoweednetlightweightautomatedweed}.}
\label{tab: EcoWeedNet_performance_SimAM_SPAB_Ground}
\resizebox{\columnwidth}{!}{%
\begin{tabular}{>{\centering\arraybackslash}m{1.5cm}cccccc}
\hline
\textbf{Model} & \textbf{Precision (\%)} & \textbf{Recall (\%)} & \textbf{mAP50 (\%)} & \textbf{mAP(50-95) (\%)} & \textbf{Parameters} & \textbf{(GFLOPs)} \\

\hline
EcoWeedNet\cite{khater2025ecoweednetlightweightautomatedweed} & \textbf{93.2} & \textbf{89} & \textbf{95.2} & \textbf{88.9} & \textbf{2.78M} & \textbf{9.3} \\
\hline
\textbf{YOLO11n\cite{ultralytics2024yolo11}}  & 89 & 88.6 & 93 & 85.6 & 2.6M & 6.5 \\ 
\hline
\textbf{YOLO12n\cite{ultralytics2025yolov12}}  & 92.8 & 84 & 93.2 & 86.9 & 2.6M & 6.5 \\ 
\hline
\textbf{YOLO4\cite{dang2023yoloweeds}}  & 94.78 & 95.04 & 95.22 & 89.48 & $\sim$66M & $\sim$141 \\
\hline
\end{tabular}%
}
\end{table}

\begin{figure}[h]
    \centering
    \setlength{\fboxsep}{0pt} 
    \setlength{\fboxrule}{0pt} 
    \includegraphics[width=\columnwidth]{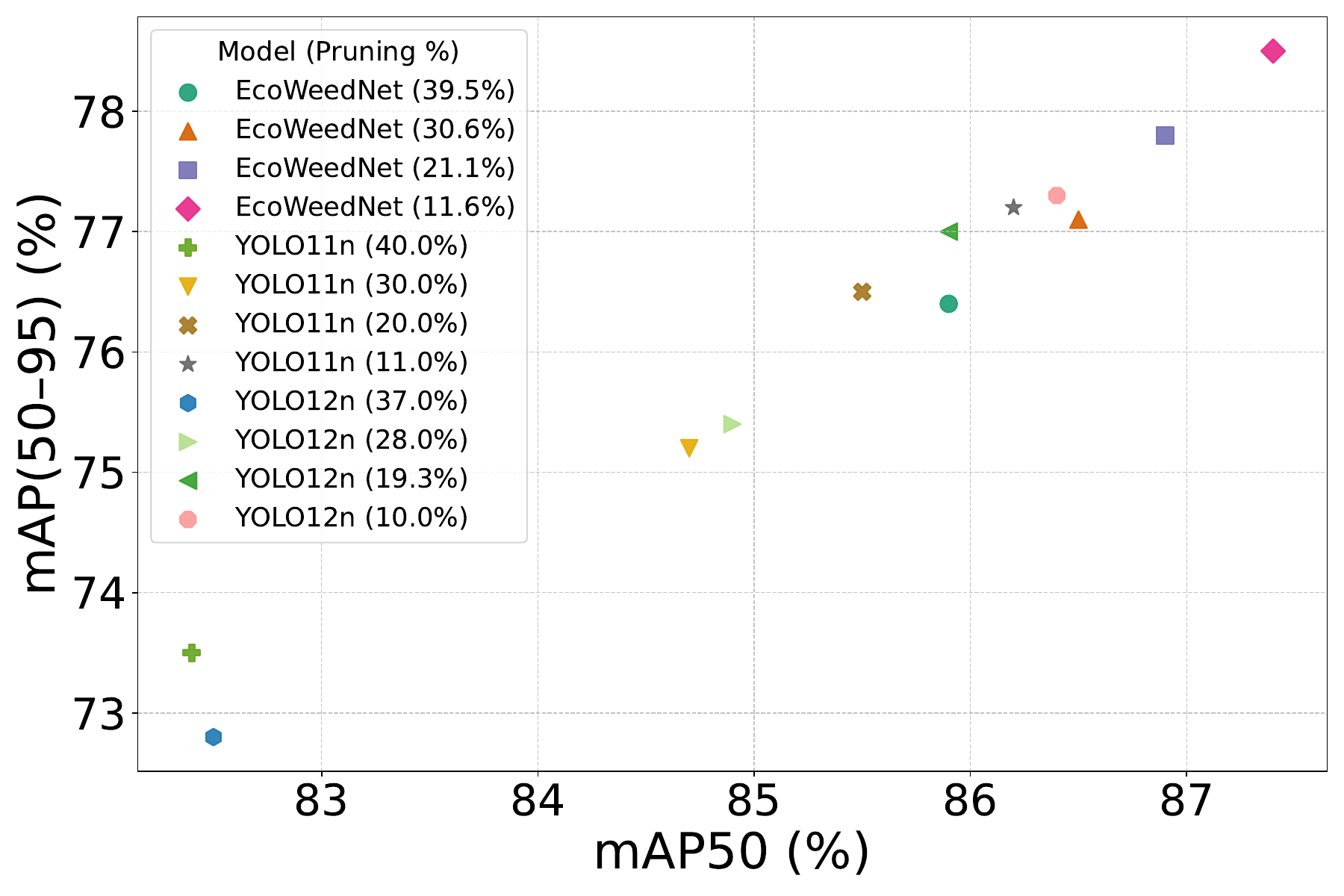}
    \caption{mAP@50 and mAP(50--95) across various pruning ratios on the CottonWeedDet12 dataset.}
    \label{fig: Ground_Performance}
\end{figure}

Table~\ref{tab: Models_Pruned_Results_Ground} and Figure~\ref{fig: Ground_Performance} present EcoWeedNet detection accuracy and the models' complexity compared to YOLO11n and YOLO12n for various channel pruning ratios. EcoWeedNet consistently proves to be the most accurate, achieving the highest evaluation scores, even after pruning a significant portion (up to 39.5\%) of its model, which has 1,683,879 parameters and 5.8 GFLOPs. It maintains a mAP50 of 85.9\%, demonstrating notably higher performance than YOLO11n, which scored 85.5\% mAP50 with 2,043,929 parameters and 5.3 GFLOPs at a pruning ratio of 20\%. YOLO12n similarly achieved a mAP50 of 85.9\% at a pruning ratio of 19.3\% with 2,072,987 parameters and 5.4 GFLOPs.

While pruning notably decreases model parameters and GFLOPs for all of them, EcoWeedNet is still able to sustain higher accuracy, showing its robustness. EcoWeedNet thus has the best overall balance of detection accuracy and minimal computation needs, therefore making it a highly promising candidate for real-world weed detection deployments, where maintaining strong accuracy with minimal computational demand is crucial.

\begin{table*}[t]
\centering
\caption{Performance of EcoWeedNet, YOLO11n, and YOLO12n under various channel pruning ratios on CottonWeedDet12 dataset.}
\label{tab: Models_Pruned_Results_Ground}
\renewcommand{\arraystretch}{1.1}
\resizebox{\textwidth}{!}{%
\begin{tabular}{c|c|cccc|c|c}
\hline
\textbf{Model} & \textbf{Pruning Ratio (\%)} & \textbf{Precision (\%)} & \textbf{Recall (\%)} & \textbf{mAP50 (\%)} & \textbf{mAP(50--95) (\%)} & \textbf{Param (M)} & \textbf{GFLOPs} \\
\hline

\multirow{4}{*}{EcoWeedNet}
 & 39.5 & 83.7 & 77.5 & 85.9 & 76.4 & 1,683,879 & 5.8 \\[2pt]
 & 30.6 & 82.1 & 80.9 & 86.5 & 77.1 & 1,931,397 & 6.6 \\[2pt]
 & 21.1 & 82.0 & 78.2 & 86.9 & 77.8 & 2,196,937 & 7.4 \\[2pt]
 & 11.6 & 84.3 & 78.1 & 87.4 & 78.5 & 2,459,176 & 8.7 \\[2pt]
\hline

\multirow{4}{*}{YOLO11} 
 & 40 & 81.7 & 73.3 & 82.4 & 73.5 & 1,564,599 & 4.2 \\[2pt]
 & 30 & 83.2 & 75.3 & 84.7 & 75.2 & 1,813,479 & 4.9 \\[2pt]
 & 20 & 77.5 & 81.4 & 85.5 & 76.5 & 2,043,929 & 5.3 \\[2pt]
 & 11 & 79.7 & 80.4 & 86.2 & 77.2 & 2,305,131 & 6.0 \\[2pt]
\hline

\multirow{4}{*}{YOLO12}
 & 37 & 78.0 & 76.0 & 82.5 & 72.8 & 1,613,675 & 4.0 \\[2pt]
 & 28 & 83.8 & 76.7 & 84.9 & 75.4 & 1,847,694 & 4.8 \\[2pt]
 & 19.3 & 80.3 & 81.5 & 85.9 & 77.0 & 2,072,987 & 5.4 \\[2pt]
 & 10 & 82.5 & 79.8 & 86.4 & 77.3 & 2,313,340 & 6.0 \\[2pt]
\hline

\end{tabular}%
}
\end{table*}




\subsection{Results After Quantization}

This section presents the outcomes obtained from the exclusive use of Quantization-Aware Training (QAT) to evaluate the performance of the EcoWeedNet and YOLO models. Our experiments focused on quantization for 32-bit and 16-bit floating-point representations across various evaluation metrics, including precision, recall, Mean Average Precision (mAP), and inference speed (FPS), using NVIDIA TensorRT on Jetson Orin Nano. The experiments utilized real-world datasets, including the Aerial Soybean Dataset and CottonWeedDet12.

\subsubsection{Results on Aerial Soybeans Dataset}

In Table~\ref{tab: Quantization_Results_Aerial}, EcoWeedNet emerges as the top performer across all measures, including precision, recall, and mAP, regardless of whether we use full 32-bit numbers or the lighter 16-bit (Float16) ones. When we switch EcoWeedNet to Float16, its accuracy drops by less than 1\%, but it runs approximately 72\% faster. In other words, the lower precision EcoWeedNet achieves the best detection accuracy and inference speed, outperforming the full-precision YOLO11n and YOLO12n, making it a wise choice for real-world precise farming.

\begin{table}[H]
\centering
\caption{Quantization results on Aerial Soybeans dataset}
\label{tab: Quantization_Results_Aerial}

\resizebox{\columnwidth}{!}{%
\begin{tabular}{c|c|cccc|c}
\hline
\textbf{Model} & \textbf{\# Bits} & \textbf{Precision (\%)} & \textbf{Recall (\%)} & \textbf{mAP50 (\%)} & \textbf{mAP(50--95) (\%)} & \textbf{TensorRT (FPS)} \\
\hline

\multirow{2}{*}{EcoWeedNet} & 32 & 71.4 & 65.7 & 72.5 & 40.8 & 83 \\
                            & 16 & 70.9 & 65.5 & 72.3 & 40.4 & 143 \\
\hline

\multirow{2}{*}{YOLO11n}    & 32 & 69.9 & 63.9 & 70.6 & 39.4 & 118 \\ 
                            & 16 & 69.6 & 63.5 & 70.5 & 39 & 175 \\
\hline

\multirow{2}{*}{YOLO12n}    & 32 & 69.7 & 62 & 70.3 & 38.1 & 77 \\
                            & 16 & 69.4 & 61.7 & 70 & 37.8 & 120 \\
\hline
\end{tabular}%
}
\end{table}

\subsubsection{Results on CottonWeedDet12 Dataset}

In Table~\ref{tab: Quantization_Results_Ground}, the performance metrics clearly indicate that the best-performing model among the compared ones is indeed EcoWeedNet. Both at 16-bit and 32-bit quantization, it performs better compared to YOLO11n and YOLO12n with higher precision, recall, and mAP. Particularly at 16-bit, it achieved a high mAP50 value of 95\% and mAP(50–95) value of 88.5\%, with robust and consistent detection performance even with reduced precision of bits. Additionally, the 16-bit EcoWeedNet version is approximately 66.7\% faster than the 32-bit version, offering competitive performance. These observations verify the efficiency of EcoWeedNet for both efficient and effective deployment under constrained resources, with an acceptable trade-off between detection accuracy and inference speed.

\begin{table}[H]
\centering
\caption{Quantization results on CottonWeedDet12 dataset}
\label{tab: Quantization_Results_Ground}

\resizebox{\columnwidth}{!}{%
\begin{tabular}{c|c|cccc|c}
\hline
\textbf{Model} & \textbf{\# Bits} & \textbf{Precision (\%)} & \textbf{Recall (\%)} & \textbf{mAP50 (\%)} & \textbf{mAP(50--95) (\%)} & \textbf{TensorRT (FPS)} \\
\hline

\multirow{2}{*}{EcoWeedNet} & 32 & 93.2 & 89 & 95.2 & 88.9 & 84 \\
                            & 16 & 92.9 & 88.8 & 95 & 88.5 & 140 \\
\hline

\multirow{2}{*}{YOLO11n}    & 32 & 89 & 88.6 & 93 & 85.6 & 112 \\ 
                            & 16 & 88.9 & 88.3 & 92.8 & 85.2 & 169 \\
\hline

\multirow{2}{*}{YOLO12n}    & 32 & 92.8 & 84 & 93.2 & 86.9 & 79 \\
                            & 16 & 92.4 & 83.7 & 93 & 86.6 & 121 \\
\hline
\end{tabular}%
}
\end{table}

\subsection{Results of Pruning and Quantization Integration}

This section includes the investigation of the impact of channel pruning and QAT integration on EcoWeedNet, YOLO11n, and YOLO12n.

\subsubsection{Results on Aerial Soybeans Dataset}

Table~\ref{tab: PQ_Results_Aerial} presents the performance of YOLO11n, YOLO12n, and EcoWeedNet after both channel pruning and quantization on the dataset of aerial soybeans. EcoWeedNet records the highest performance on precision, recall, and average mean precision (mAP) even after pruning 11.6\% of its channels, the highest among the models' pruning ratios. 

Quantization of EcoWeedNet to Float16 affects its accuracy by a negligible decrease of less than 1\%, while it accelerates inference by more than 77\%. This reflects the robustness of EcoWeedNet to aggressive compression, with no impact on detection accuracy and a notable improvement in inference speeds, making it suitable for deployment in real-world farming applications.

\begin{table}[h]
\centering
\caption{(Pruning + Quantization) results on Aerial Soybeans dataset}
\label{tab: PQ_Results_Aerial}

\resizebox{\columnwidth}{!}{%
\begin{tabular}{c|c|c|cccc|c}
\hline
\textbf{Model} & \textbf{Pruning Ratio} & \textbf{\# Bits} & \textbf{Precision (\%)} & \textbf{Recall (\%)} & \textbf{mAP50 (\%)} & \textbf{mAP(50--95) (\%)} & \textbf{TensorRT (FPS)} \\
\hline

\multirow{2}{*}{EcoWeedNet} & \multirow{2}{*}{11.6 \%} & 32 &  64.5 & 56.4 & 63.4 & 34.3 & 104 \\
                            &  & 16 & 64.2 & 56.2 & 63 & 33.9 & 184 \\
\hline
\multirow{2}{*}{YOLO11n}    & \multirow{2}{*}{11 \%} & 32 & 60.6 & 55.3 & 60.4 & 31.4 & 129 \\ 
                            &  & 16 & 60.3 & 55.1 & 60.2 & 31.1 & 194 \\
\hline
\multirow{2}{*}{YOLO12n}    & \multirow{2}{*}{10 \%} & 32 & 61.9 & 51.6 & 60.7 & 31.7 & 95 \\
                            &  & 16 & 61.6 & 51.2 & 60.5 & 31.4 & 150 \\
\hline
\end{tabular}%
}
\end{table}

\subsubsection{Results on CottonWeedDet12 Dataset}

As shown in Table~\ref{tab:PQ_Results_Ground}, the integration of quantization and pruning for the CottonWeedDet12 dataset provides strong support for EcoWeedNet as the optimal choice. With a pruning ratio of 11.6\% with 16-bit quantization, the best performance values are achieved by EcoWeedNet with an mAP50 of 87.1\% and an mAP(50–95) of 78.1\% over the benchmarking YOLO11n and YOLO12n under these constraints. We further observed that the 16-bit version of EcoWeedNet achieved a high speed of 180 FPS, surpassing YOLO12n and comparable to YOLO11n. This demonstrates the robustness of EcoWeedNet in maintaining precision even under quantization and compression scenarios, and thus proves to be suitable for real-time object detection tasks in systems with limited resources.

\begin{table}[ht]
\centering
\caption{(Pruning + Quantization) results on CottonWeedDet12 dataset}
\label{tab:PQ_Results_Ground}

\resizebox{\columnwidth}{!}{%
\begin{tabular}{c|c|c|cccc|c}
\hline
\textbf{Model} & \textbf{Pruning Ratio} & \textbf{\# Bits} & \textbf{Precision (\%)} & \textbf{Recall (\%)} & \textbf{mAP50 (\%)} & \textbf{mAP(50--95) (\%)} & \textbf{TensorRT (FPS)} \\
\hline

\multirow{2}{*}{EcoWeedNet} & \multirow{2}{*}{11.6 \%} &32 &  84.3 & 78.1 & 87.4 & 78.5 & 105 \\
&  & 16 & 84 & 77.8 & 87.2 & 78.1 & 180  \\
\hline
\multirow{2}{*}{YOLO11n}    & \multirow{2}{*}{11 \%} & 32 & 79.7 & 80.4 & 86.2 & 77.2 & 131 \\ 
&  & 16 & 79.7 & 80 & 86 & 76.8 & 189 \\
\hline
\multirow{2}{*}{YOLO12n}    & \multirow{2}{*}{10 \%} & 32 &  82.5 & 79.8 & 86.4 & 77.3 & 95 \\
&  & 16 & 82.1 & 79.6 & 86 & 77 & 146 \\
\hline
\end{tabular}%
}
\end{table}

\begin{table*}[t]
\centering
\caption{GPU Efficiency Analysis}
\label{tab: gpu_performance_Analysis}
\resizebox{\textwidth}{!}{%
\begin{tabular}{c|c|c|c|c|c|c|c|c|c}
\hline
\textbf{Model} & \textbf{Precision} & \textbf{Pruning Ratio (\%)} & \textbf{Throughput (QPS)} & \textbf{Host Mean Latency (ms)} & \textbf{GPU Compute Mean (ms)} & \textbf{Coeff.\ of Var.\ (\%)} & \textbf{GPU Utilization (\%)} & \textbf{Ave. Power Draw (W)} & \textbf{Peak Power Draw (W)} \\ [2pt]
\hline 
\multirow{4}{*}{EcoWeedNet} 
& FP32 & 0 & 79 & 13.02 & 12.6 & 20.6 & 89.29 & 15.79 & 16.16 \\ 
& FP16 & 0 & 166.5 & 6.35 & 5.97 & 10.3 & 87.75 & 13.73 & 15.20 \\ 
& FP32 & 11.6 & 114.5 & 9.04 & 8.73 & 9.47 & 88.60 & 15.33 & 15.64 \\ 
& FP16 & 11.6 & 225.038 & 4.81 & 4.43 & 21.29 & 82.67 & 11.58 & 13.88 \\ 

\hline

\multirow{4}{*}{YOLO11n} 
& FP32 & 0 & 122.2 & 8.585 & 8.155 & 21.2 & 78.50 & 14.41 & 15.64 \\ 
& FP16 & 0 & 241.67 & 5.10 & 4.52 & 8.25 & 73.00 & 11.73 & 13.72 \\ 
& FP32 & 11 & 177.19 & 5.92 & 5.63 & 9.75 & 87.56 & 13.98 &  14.89\\ 
& FP16 & 11 & 296.51 & 3.75 & 3.36 & 15.08 & 88.05 & 9.95 & 13.23 \\ 

\hline

\multirow{4}{*}{YOLO12n} 
& FP32 & 0 & 79.82 & 12.79 & 12.47 & 9.26 & 90.43 & 15.70 &  16.28 \\ 
& FP16 & 0 & 137.8 & 7.61 & 7.23 & 6.99 & 85.17 & 12.89 &  14.59 \\ 
& FP32 & 10 & 103.73 & 8.83 & 8.61 & 4.26 & 90.23 & 15.24 & 15.23  \\ 
& FP16 & 10 & 204.95 & 5.56 & 4.75 & 13.97 & 88.69 & 11.31 &  14.22 \\ 

\hline
\end{tabular}%
}
\end{table*}

\subsection{Memory Consumption Analysis}

Table~\ref{tab: memory_comparison} shows that switching from FP32 to FP16 lowers memory requirements across all three models.  The engine size on disk for YOLO11 shrinks by 30\,\%, YOLO12 by 31\,\%, and EcoWeedNet by 45\,\%.  

More importantly for deployment, the device scratch memory also drops: YOLO11 by~49\,\%, YOLO12 by~50\,\%, and EcoWeedNet by~46\,\%. A similar trend appears in the execution‐context memory, with reductions of 50\,\% for YOLO11, 50\,\% for YOLO12, and 47\,\% for EcoWeedNet.  

Averaged over all models, FP16 cuts the engine footprint by $\sim$35\,\%, scratch memory by $\sim$49\,\%, and context memory by $\sim$49\,\%.  
These savings confirm that half‐precision reduces storage and lowers the peak GPU memory footprint, making FP16 engines better suited for memory-constrained edge devices.

In the FP16 configuration, EcoWeedNet requires only 7.2\,MB for its TensorRT engine file, making it 1.6\,MB and 2.0\,MB smaller than the YOLO11n and YOLO12n engines, respectively. This compact engine size offers several practical advantages, including reduced storage requirements. According to NVIDIA, the engine size also closely approximates the amount of GPU memory needed to store model weights during deserialization~\cite{nvidia_how_trt_works}, so a smaller engine size is better. 

\begin{table}[H]
\centering
\caption{Memory Consumption Comparison in FP32 and FP16 Precision}
\label{tab: memory_comparison}
\resizebox{\columnwidth}{!}{%
\begin{tabular}{c|c|c|c|c}
\hline
\textbf{Model} & \textbf{Precision} & \textbf{Engine Size (MB)} & \textbf{Device Scratch (MB)} & \textbf{Context Memory (MB)} \\
\hline
\multirow{2}{*}{EcoWeedNet}& FP32 & 13.0 & 22.80 & 21.88 \\
 & FP16 & 7.2 & 12.23 & 11.67 \\
 \hline

\multirow{2}{*}{YOLO11n}   & FP32 & 12.5 & 18.85 & 17.97 \\
 & FP16 & 8.8 & 9.53 & 9.08 \\
 \hline

\multirow{2}{*}{YOLO12n}   & FP32 & 13.4 & 24.16 & 23.05 \\
 & FP16 & 9.2 & 12.08 & 11.52 \\
\hline
\end{tabular}%
}
\end{table}

\subsection{GPU Efficiency Analysis}


As shown in Table~\ref{tab: gpu_performance_Analysis}, switching from full precision (FP32) to half-precision (FP16) offers an improvement in GPU efficiency in terms of latency, power draw, and throughput (Queries Per Second), which measures the number of requests for information processed per second. The average power draw was computed over ten inference runs by first averaging the instantaneous readings within each run and then taking the mean of those run-level averages. All other reported values are simple means over the ten runs.

FP32 version of EcoWeedNet’s throughput jumps from 79 to 166.5\,QPS, which is about a 111\% increase, while its GPU-side latency falls by roughly 53\% (12.6\,ms to 5.97\,ms), and average power drops by 13\%. YOLO11n achieves almost double the throughput, from 122.2 to 241.67\,QPS, along with a 45\% latency decrease and a 19\% reduction in power draw. In terms of power draw, YOLO12n consumes 18\% less power, but gains the smallest improvement in throughput with only a 73\% increase, and a 42\% reduction in latency. In conclusion, quantization demonstrates its impact in making networks faster during inference and more battery-friendly.

Combining pruning and quantization has an even greater impact on GPU efficiency. Comparison of the full-precision baseline models with the pruned and quantized models shows that EcoWeedNet increases its throughput from 79 to 225 QPS, achieving a 185\% increase. At the same time, latency drops by 65\% and power consumption decreases by 27\%. YOLO11n experiences a 143\% boost in throughput, a 59\% latency reduction, and a 31\% drop in power draw. YOLO12n improves by 157\% in throughput, with a 62\% decrease in latency and a 28\% reduction in power consumption. Overall, pruning and quantization integrate their benefits, significantly enhancing inference speed, reducing power consumption, and increasing suitability for real-time deployment.

Comparing the FP16 version of EcoWeedNet with the FP32 YOLO11n and YOLO12n. The proposed EcoWeedNet achieves the best results in terms of GPU efficiency, including the highest throughput, lowest latency, and lowest average power draw. Additionally, as shown in Table~\ref{tab: Quantization_Results_Ground} and Table~\ref{tab: Quantization_Results_Aerial}, the quantized EcoWeedNet achieves the highest scores in terms of mean average precision, recall, and precision. Moreover, as shown in Table~\ref{tab:PQ_Results_Ground} and Table~\ref{tab: PQ_Results_Aerial}, the pruned and quantized EcoWeedNet also achieves the best results across most detection metrics. Consequently, EcoWeedNet is the optimal choice for deployment in energy and memory-constrained edge devices.

\subsection{Layer-wise GPU Latency Analysis.}

Table~\ref{tab: gpu_latency_key} shows the most time-consuming convolutional layer in the three networks. EcoWeedNet’s bottleneck, a convolutional layer, lasts $0.4076$ ms, accounting for $3.4\%$ of total inference time. YOLO11n takes $0.2033$ ms with $2.4\%$ of total inference time, while YOLO12n’s first convolutional layer requires $0.3636$ ms, contributing $2.8\%$ of total inference. This indicates that no single layer causes a significant delay, suggesting that no specific layer needs a specific optimization. 

Although the layer itself is structurally the same in YOLO11n and YOLO12n, the latency differs due to architectural differences, which influence runtime scheduling, memory access patterns, and resource allocation, even when using the same hardware platform and software environment~\cite{mendoza2020predicting}.

\begin{table}[h]
\centering
\caption{Per-layer GPU latency Comparison}
\label{tab: gpu_latency_key}
\resizebox{\columnwidth}{!}{%
\begin{tabular}{cccc}
\hline
\textbf{Model} & \textbf{Layer (Time)} & \textbf{Avg Time (ms)} & \textbf{\% Time} \\ [2pt]
\hline

EcoWeedNet & /model.3/c1\_r/eval\_conv/Conv & 0.4076 & 3.4 \\ 

YOLO11n & /model.0/conv/Conv & 0.2033 & 2.4  \\ 

YOLO12n & /model.0/conv/Conv & 0.3636 & 2.8 \\ 
\hline
\end{tabular}%
}
\end{table}

\subsection{Discussion and Insights}

Experimental results convincingly demonstrate that the optimized EcoWeedNet is optimal for effective and efficient weed detection in computationally constrained edge devices. Across all compression scenarios involving structured channel pruning, quantization-aware training (QAT), and their integration, EcoWeedNet consistently outperformed YOLO11n and YOLO12n in detection performance. Most importantly, the model remains robust, maintaining high mean Average Precision (mAP) values even when computational costs and memory utilization were notably reduced. 

Furthermore, EcoWeedNet exhibited notable adaptability, recovering quickly from the pruned state due to the fine-tuning till convergence, where the pruning operation occurs after convergence, ensuring the highest accuracy is preserved. The experiment highlights EcoWeedNet's potential for real-time applications in precision farming, where model performance must be balanced with computational costs.

Combining pruning and quantization can cause cumulative approximation errors, potentially degrading model accuracy. To address this, we used quantization-aware training (QAT), enabling the model to adapt to low-precision noise during training. Additionally, pruning was performed in a structured, channel-wise manner based on importance metrics, ensuring minimal impact on critical features. Fine-tuning was done after pruning to recover the lost detection accuracy. This staged process helps prevent significant decreases in detection accuracy caused by quantization and pruning.

\section{Conclusion and Future Work}

In this study, we demonstrate the effectiveness of the structured pruning, quantization-aware training (QAT) method, and the integration between them approach in optimizing EcoWeedNet for real-time weed detection on resource-constrained edge devices. Our pruning approach addresses the inherent complexity of YOLO modules while maintaining their consistency in the architecture after pruning. With extensive evaluation using two datasets and varying pruning ratios, the compressed EcoWeedNet outperformed the state-of-the-art YOLO models, demonstrating its robustness and applicability in real-world scenarios. The results indicate that the compressed version of EcoWeedNet is an efficient and scalable solution for practical agricultural deployment, offering accelerated weed detection capability using NVIDIA Jetson Orin Nano for precision agriculture.

Future work involves investigating novel approaches to calculating the importance of channels in both spatial and frequency domains to select the less informative channels more accurately and minimize losses in detection accuracy after pruning. Additionally, mixed-precision quantization methods require further investigation to optimize model efficiency even more.


\section*{Acknowledgement}
Authors acknowledge the support received from the Deanship of Research, King Fahd University of Petroleum \& Minerals (KFUPM), and SDAIA-KFUPM Joint Research Center for Artificial Intelligence through Grant\# JRC-AI-RFP-17 (CAI02562)

\bibliographystyle{IEEEtran}
\bibliography{references}

\begin{thebibliography}{10}
\providecommand{\url}[1]{#1}
\csname url@samestyle\endcsname
\providecommand{\newblock}{\relax}
\providecommand{\bibinfo}[2]{#2}
\providecommand{\BIBentrySTDinterwordspacing}{\spaceskip=0pt\relax}
\providecommand{\BIBentryALTinterwordstretchfactor}{4}
\providecommand{\BIBentryALTinterwordspacing}{\spaceskip=\fontdimen2\font plus
\BIBentryALTinterwordstretchfactor\fontdimen3\font minus \fontdimen4\font\relax}
\providecommand{\BIBforeignlanguage}[2]{{%
\expandafter\ifx\csname l@#1\endcsname\relax
\typeout{** WARNING: IEEEtran.bst: No hyphenation pattern has been}%
\typeout{** loaded for the language `#1'. Using the pattern for}%
\typeout{** the default language instead.}%
\else
\language=\csname l@#1\endcsname
\fi
#2}}
\providecommand{\BIBdecl}{\relax}
\BIBdecl

\bibitem{saikumar2024drive}
D.~Saikumar and B.~Varghese, ``Drive: Dual gradient-based rapid iterative pruning,'' \emph{arXiv preprint arXiv:2404.03687}, 2024.

\bibitem{mallya2018packnet}
A.~Mallya and S.~Lazebnik, ``Packnet: Adding multiple tasks to a single network by iterative pruning,'' in \emph{Proceedings of the IEEE conference on Computer Vision and Pattern Recognition}, 2018, pp. 7765--7773.

\bibitem{jacob2018quantization}
B.~Jacob, S.~Kligys, B.~Chen, M.~Zhu, M.~Tang, A.~Howard, H.~Adam, and D.~Kalenichenko, ``Quantization and training of neural networks for efficient integer-arithmetic-only inference,'' in \emph{Proceedings of the IEEE conference on computer vision and pattern recognition}, 2018, pp. 2704--2713.

\bibitem{10536014}
J.~Lee, Y.~Kwon, S.~Park, M.~Yu, J.~Park, and H.~Song, ``Q-hyvit: Post-training quantization of hybrid vision transformers with bridge block reconstruction for iot systems,'' \emph{IEEE Internet of Things Journal}, vol.~11, no.~22, pp. 36\,384--36\,396, 2024.

\bibitem{9367271}
K.~Zhang, H.~Ying, H.-N. Dai, L.~Li, Y.~Peng, K.~Guo, and H.~Yu, ``Compacting deep neural networks for internet of things: Methods and applications,'' \emph{IEEE Internet of Things Journal}, vol.~8, no.~15, pp. 11\,935--11\,959, 2021.

\bibitem{kim2021pqk}
J.~Kim, S.~Chang, and N.~Kwak, ``Pqk: model compression via pruning, quantization, and knowledge distillation,'' \emph{arXiv preprint arXiv:2106.14681}, 2021.

\bibitem{10965714}
J.~Lu, G.~Bao, X.~Deng, X.~Han, Y.~Lan, and H.~Wu, ``Iot-based precision litchi tracking and counting method using gated metrics,'' \emph{IEEE Internet of Things Journal}, pp. 1--1, 2025.

\bibitem{khater2025ecoweednetlightweightautomatedweed}
O.~H. Khater, A.~J. Siddiqui, M.~S. Hossain, and A.~El-Maleh, ``Ecoweednet: A lightweight and automated weed detection method for sustainable next-generation agricultural consumer electronics,'' \emph{IEEE Transactions on Consumer Electronics}, pp. 1--1, 2025.

\bibitem{ultralytics2024yolo11}
Ultralytics, ``{YOLOv11}: Next-generation object detection model,'' \url{https://docs.ultralytics.com/models/yolo11/}, 2024, accessed: 2025-04-20.

\bibitem{ultralytics2025yolov12}
\BIBentryALTinterwordspacing
------, ``Yolov12 - ultralytics documentation,'' 2024, accessed: 2025-04-10. [Online]. Available: \url{https://docs.ultralytics.com/models/yolo12/}
\BIBentrySTDinterwordspacing

\bibitem{li2022revisiting}
Y.~Li, K.~Adamczewski, W.~Li, S.~Gu, R.~Timofte, and L.~Van~Gool, ``Revisiting random channel pruning for neural network compression,'' in \emph{Proceedings of the IEEE/CVF conference on computer vision and pattern recognition}, 2022, pp. 191--201.

\bibitem{lin2020channel}
M.~Lin, R.~Ji, Y.~Zhang, B.~Zhang, Y.~Wu, and Y.~Tian, ``Channel pruning via automatic structure search,'' \emph{arXiv preprint arXiv:2001.08565}, 2020.

\bibitem{mondal2022adaptive}
M.~Mondal, B.~Das, S.~D. Roy, P.~Singh, B.~Lall, and S.~D. Joshi, ``Adaptive cnn filter pruning using global importance metric,'' \emph{Computer Vision and Image Understanding}, vol. 222, p. 103511, 2022.

\bibitem{yang2022channel}
C.~Yang and H.~Liu, ``Channel pruning based on convolutional neural network sensitivity,'' \emph{Neurocomputing}, vol. 507, pp. 97--106, 2022.

\bibitem{zhang2022carrying}
Y.~Zhang, M.~Lin, C.-W. Lin, J.~Chen, Y.~Wu, Y.~Tian, and R.~Ji, ``Carrying out cnn channel pruning in a white box,'' \emph{IEEE Transactions on Neural Networks and Learning Systems}, vol.~34, no.~10, pp. 7946--7955, 2022.

\bibitem{liu2023eacp}
Y.~Liu, D.~Wu, W.~Zhou, K.~Fan, and Z.~Zhou, ``Eacp: An effective automatic channel pruning for neural networks,'' \emph{Neurocomputing}, vol. 526, pp. 131--142, 2023.

\bibitem{geng2024complex}
X.~Geng, J.~Gao, Y.~Zhang, and D.~Xu, ``Complex hybrid weighted pruning method for accelerating convolutional neural networks,'' \emph{Scientific Reports}, vol.~14, no.~1, p. 5570, 2024.

\bibitem{zhu2025comprehensive}
K.~Zhu, F.~Hu, Y.~Ding, W.~Zhou, and R.~Wang, ``A comprehensive review of network pruning based on pruning granularity and pruning time perspectives,'' \emph{Neurocomputing}, p. 129382, 2025.

\bibitem{9714269}
S.~Liu, L.~T. Yang, X.~Tu, R.~Li, and C.~Xu, ``Lightweight monocular depth estimation on edge devices,'' \emph{IEEE Internet of Things Journal}, vol.~9, no.~17, pp. 16\,168--16\,180, 2022.

\bibitem{liu2024mpq}
X.~Liu, T.~Wang, J.~Yang, C.~Tang, and J.~Lv, ``Mpq-yolo: Ultra low mixed-precision quantization of yolo for edge devices deployment,'' \emph{Neurocomputing}, vol. 574, p. 127210, 2024.

\bibitem{idama2024qatfp}
G.~Idama, Y.~Guo, and W.~Yu, ``Qatfp-yolo: Optimizing object detection on non-gpu devices with yolo using quantization-aware training and filter pruning,'' in \emph{2024 33rd International Conference on Computer Communications and Networks (ICCCN)}.\hskip 1em plus 0.5em minus 0.4em\relax IEEE, 2024, pp. 1--6.

\bibitem{moosmann2023flexible}
J.~Moosmann, H.~Mueller, N.~Zimmerman, G.~Rutishauser, L.~Benini, and M.~Magno, ``Flexible and fully quantized ultra-lightweight tinyissimoyolo for ultra-low-power edge systems,'' \emph{arXiv preprint arXiv:2307.05999}, 2023.

\bibitem{xue2024yolo}
C.~Xue, Y.~Xia, M.~Wu, Z.~Chen, F.~Cheng, and L.~Yun, ``El-yolo: An efficient and lightweight low-altitude aerial objects detector for onboard applications,'' \emph{Expert Systems with Applications}, vol. 256, p. 124848, 2024.

\bibitem{dong2022pg}
C.~Dong, C.~Pang, Z.~Li, X.~Zeng, and X.~Hu, ``Pg-yolo: A novel lightweight object detection method for edge devices in industrial internet of things,'' \emph{IEEE Access}, vol.~10, pp. 123\,736--123\,745, 2022.

\bibitem{zhang2024robust}
X.~Zhang, Y.~Feng, S.~Zhang, N.~Wang, G.~Lu, and S.~Mei, ``Robust aerial person detection with lightweight distillation network for edge deployment,'' \emph{IEEE Transactions on Geoscience and Remote Sensing}, 2024.

\bibitem{su2024yolic}
K.~Su, Y.~Tomioka, Q.~Zhao, and Y.~Liu, ``Yolic: An efficient method for object localization and classification on edge devices,'' \emph{Image and Vision Computing}, vol. 147, p. 105095, 2024.

\bibitem{10891155}
Y.~Wang, S.~Liu, B.~Guo, B.~Zhang, K.~Ma, Y.~Ding, H.~Luo, Y.~Li, and Z.~Yu, ``Adascale: Dynamic context-aware dnn scaling via automated adaptation loop on mobile devices,'' \emph{IEEE Internet of Things Journal}, pp. 1--1, 2025.

\bibitem{lu2023cottonweeddet12}
\BIBentryALTinterwordspacing
Y.~Lu, ``Cottonweeddet12: a 12-class weed dataset of cotton production systems for benchmarking ai models for weed detection,'' 2023. [Online]. Available: \url{https://weed-ai.sydney.edu.au/datasets/2c14915b-0827-4b65-9908-d2a6df0d48f3}
\BIBentrySTDinterwordspacing

\bibitem{rehman2024agriweed}
M.~U. Rehman, ``Agriweeddetection,'' \url{https://github.com/Rehman1995/AgriWeedDetection}, 2024, gitHub repository, accessed: 2025-04-22.

\bibitem{10948450}
L.~Nkenyereye, C.~Rajkumar, B.~G. Lee, and W.-Y. Chung, ``Dynamic transfer learning switching approach using resource benchmark in edge intelligence,'' \emph{IEEE Internet of Things Journal}, pp. 1--1, 2025.

\bibitem{yolotrt2022}
J.~Lin, ``Yolotrt: tensorrt for yolo series,'' \url{[https://github.com/Linaom1214/TensorRT-For-YOLO-Series]}, 2022.

\bibitem{dang2023yoloweeds}
F.~Dang, D.~Chen, Y.~Lu, and Z.~Li, ``Yoloweeds: A novel benchmark of yolo object detectors for multi-class weed detection in cotton production systems,'' \emph{Computers and Electronics in Agriculture}, vol. 205, p. 107655, 2023.

\bibitem{nvidia_how_trt_works}
\BIBentryALTinterwordspacing
{NVIDIA Corporation}, ``{How TensorRT Works},'' 2025, accessed: 2025-06-15. [Online]. Available: \url{https://docs.nvidia.com/deeplearning/tensorrt/latest/architecture/how-trt-works.html}
\BIBentrySTDinterwordspacing

\bibitem{mendoza2020predicting}
D.~M. Mendoza and S.~Wang, ``Predicting latency of neural network inference,'' 2020.

\end{thebibliography}

\end{document}